\documentclass[10pt,twocolumn,letterpaper]{article}

\usepackage{iccv}
\usepackage{times}
\usepackage{epsfig}
\usepackage{graphicx}
\usepackage{amsmath}
\usepackage{amssymb}
\usepackage{subfigure}
\usepackage{multirow}
\usepackage[pagebackref=true,breaklinks=true,letterpaper=true,colorlinks,bookmarks=false]{hyperref}

\iccvfinalcopy % *** Uncomment this line for the final submission

\def\iccvPaperID{9711} % *** Enter the ICCV Paper ID here

\ificcvfinal\fi

\begin{document}

%%%%%%%%% TITLE
\title{MixDehazeNet : Mix Structure Block For Image Dehazing Network}

\author{LiPing Lu\textsuperscript{*}\\
WuHan University Of Technology\\
% % First line of institution2 address\\
{\tt\small luliping@whut.edu.cn}
\and
Qian Xiong\thanks{Equal contribution.}\\
WuHan University Of Technology\\
{\tt\small xiongqian2021@whut.edu.cn}
\and
DuanFeng Chu\\
WuHan University Of Technology\\
% % First line of institution2 address\\
{\tt\small chudf@whut.edu.cn}
\and
BingRong Xu\thanks{Corresponding author}\\
WuHan University Of Technology\\
% % First line of institution2 address\\
{\tt\small bingrongxu@whut.edu.cn}
}

\maketitle
\ificcvfinal\thispagestyle{empty}\fi

%%%%%%%%% ABSTRACT
\begin{abstract}
Image dehazing is a typical task in the low-level vision field. Previous studies verified the effectiveness of the large convolutional kernel and attention mechanism in dehazing. However, there are two drawbacks: the multi-scale properties of an image are readily ignored when a large convolutional kernel is introduced, and the standard series connection of an attention module does not sufficiently consider an uneven hazy distribution. In this paper, we propose a novel framework named Mix Structure Image Dehazing Network (MixDehazeNet), which solves two issues mentioned above. Specifically, it mainly consists of two parts: the multi-scale parallel large convolution kernel module and the enhanced parallel attention module. Compared with a single large kernel, parallel large kernels with multi-scale are more capable of taking partial texture into account during the dehazing phase. In addition, an enhanced parallel attention module is developed, in which parallel connections of attention perform better at dehazing uneven hazy distribution. Extensive experiments on three benchmarks demonstrate the effectiveness of our proposed methods. For example, compared with the previous state-of-the-art methods, MixDehazeNet achieves a significant improvement (42.62dB PSNR) on the SOTS indoor dataset.  The code is released in \url{https://github.com/AmeryXiong/MixDehazeNet}.

\end{abstract}

%%%%%%%%% BODY TEXT
%-------------------------------------------------------------------------
\section{Introduction}

\begin{figure}[t]
\begin{center}
    \includegraphics[width=1\linewidth]{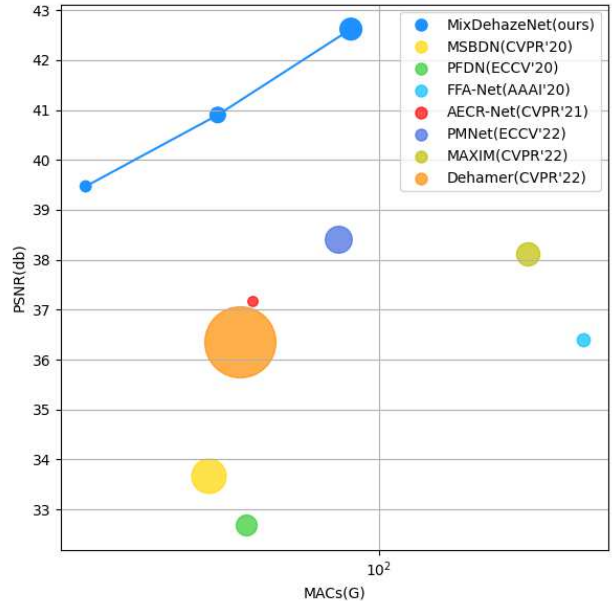}
\end{center}
   \caption{The results of MixDehazeNet compared with other dehazing methods on SOTS indoor dataset.The size of the circle represents \#Param, and MACs are shown with the logarithmic axis.}
   \label{Figure 1}
\label{fig:long}
\label{fig:onecol}
\end{figure}

Image dehazing is an important low-level task in computer vision. Haze generally exists in images, videos, and other visual scenarios, which degrades human recognition. The majority of computer vision tasks such as target detection \cite{yolov3,maskRcnn,NAS-FPN}, re-recognition \cite{LifelongPersonRe-identification,GeneralizablePersonRe-identification}, and semantic segmentation \cite{FCNSegmentation,DeeplabSegmentation,PyramidSegmentation} are implicitly influenced by hazy images and videos, which reduces the performance of the deep models. %Therefore, image dehazing is essential for computer high-level tasks, which benefits image and video analysis and further increase the stability of aforementioned tasks. 
%Therefore, image dehazing is essential for computer high level tasks, it can increase the stability of computer high level tasks, but also can bring clear hazy-free images to people. 
Therefore, single-image dehazing which aims to recover the clear scene from the matching hazy image has attracted great attention among both academic and industrial communities. It can serve as the first step in the preprocessing stage of the following high-level vision tasks as a fundamental low-level picture restoration task.

% Haze often occurs in people's daily lives. Haze can reduce the human eye's recognition. It can also reduce the accuracy of computer high level tasks, such as , which will damage the auto-driving system and traffic safety monitoring system, and reduce human safety. 

% Early methods for image dehazing \cite{berman2016non,fattal2014dehazing,he2010single,zhu2015fast} focus on estimating the construction function of haze. 

The goal of image dehazing is to restore a hazy image to a haze-free image. Atmospheric scattering models \cite{mccartney1976optics,narasimhan2002vision,nayar1999vision} are commonly used to explain the process of image dehazing. Formally, given an image $x$, $J(x)$ represents its haze-free image vision. The hazy image can be written as:
\begin{equation}
I(x)=J(x)t(x)+A(1-t(x)),
\end{equation}
where $A$ is the global atmospheric light and $t(x)$ is the medium transmission map. Further, $t(x)$ can be formulated as $t(x)=e^{-\beta}d(x)$, where ${\beta}$ is the scattering coefficient of the atmosphere and $d(x)$ is the scene depth. 
Early image dehazing methods \cite{berman2016non,fattal2014dehazing,he2010single,zhu2015fast} were based on prior knowledge, and prior knowledge was used to estimate $A$ and $t(x)$. While these methods performed well under prior assumptions, the recovered image could be distorted if the prior assumptions were not met.

% Early image dehazing methods \cite{berman2016non,fattal2014dehazing,he2010single,zhu2015fast} heavily relied on prior knowledge, \textit{i.e.}, estimating $A$ and $t(x)$ via prior knowledge. These methods worked effectively when a certain presumption was valid, but the reconstructed image would be degraded otherwise.

The development of deep learning has led to significant advancements in image dehazing. Existing dehazing methods can be broadly classified into two categories based on network architecture: 1) CNN-based methods \cite{GCANet,qin2020ffa,LKD-net}, which mostly focus on increasing the depth and width of the networks or using large convolution kernels. Large convolution kernels\cite{VAN,largeKernel} can capture more structured information in the learned latent domain space by expanding their receptive field. 2) Transformer-based methods \cite{Dehamer,valanarasu2022transweather,ji2021u2}, which have global modeling capabilities and large receptive fields, but requiring a large number of parameters and a huge-cost training process.

Despite remarkable performance of current methods, there are two limitations: 1) Although both CNN-based and Transformer-based methods can leverage large effective receptive fields to enhance performance, the multi-scale characteristics of images are always ignored during the dehazing processes. The haze concentration distribution in each image is non-uniform, and different sizes of convolution kernels can effectively capture haze distribution areas of different scales. 2) The attention mechanism used in previous dehazing networks\cite{qin2020ffa,AECR-net,GCANet} was not entirely suitable for image dehazing. We notice that channel attention might be able to better encode global share variable $A$ while pixel attention might be able to better encode location-dependent local variable $t(x)$. But existing methods\cite{qin2020ffa,AECR-net,GCANet} only designed the pixel attention module and channel attention module separately.

To address these issues, %we propose a novel Mix Structure Block Dehazing Network named MixDehazeNet for image dehazing. The core point of MixDehazeNet is the mix structure block, which can be integrated into U-net \cite{u-net} for enhancing efficient dehazing. 
we propose a novel Mix Structure Block Dehazing Network named MixDehazeNet, for image dehazing, which utilizes the U-net \cite{u-net} as the backbone and contains mix structure block which combines multi-scale parallel large convolution kernel module and enhanced parallel attention module. Mix structure block is a transformer-style block that replaces the multi-head self-attention in the transformer with a multi-scale parallel large convolution kernel module, and replaces the feed-forward network in the transformer with an enhanced parallel attention module.
%Specifically, the multi-scale parallel large convolution kernel module and the enhanced parallel attention module are two components of the mix structure block, which is a transformer-style block. 
%Firstly, we introduce 
The multi-scale parallel large convolution kernel module, called MSPLCK, has both multi-scale characteristics and large receptive fields. In this module, the large convolution concentrates on global features and catches regions with significant haze, while the small convolution concentrates on detailed features and restores texture details.  Also, we design the enhanced parallel attention module called EPA, which has the ability to jointly employ channel attention to extract shared global information of the original feature and pixel attention to extract location-dependent local information of the original feature in parallel, making it capable of dealing with an uneven hazy distribution effectively. This module includes three attention mechanisms (simple pixel attention, channel attention, and pixel attention), which are fused through a multi-layer perception. Moreover, inspired by AECR-Net \cite{AECR-net}, the contrastive loss is integrated with the proposed model to enhance performance. Different from AECR-Net, MixDehazeNet uses ResNet-152 \cite{ResNet} as a backbone for contrastive learning as we found it to be more effective than VGG19 \cite{Vgg} in improving the performance of our model. Our contribution can be summarized as the follows:

\begin{itemize}
    \item We designed the multi-scale parallel large convolution kernel module with large receptive fields and multi-scale properties. It can simultaneously recover texture details while capturing large hazed areas. The parallel dilated convolution also has large receptive fields and long-distance modeling capability.
\end{itemize}

 \begin{itemize}
    \item  We designed the enhanced parallel attention module that can efficiently deal with uneven hazy distribution and is more suitable for image dehazing. It can extract shared global information and location-dependent local information of the original feature in parallel.
\end{itemize}

\begin{itemize}
    \item %we propose a novel Mix Structure Block Dehazing Network named MixDehazeNet for image dehazing. Figure \ref{Figure 1} shows MixDehazeNet compares with other SOTA models in SOTS indoor dataset. 
    Overall, the proposed MixDehazeNet have achieved state-of-the-art results on multiple image dehazing datasets. Figure \ref{Figure 1} shows MixDehazeNet compares with other SOTA models in SOTS indoor dataset. To our best knowledge, MixDehazeNet-L is the first model to exceed 42dB PSNR on the SOTS indoor dataset.
\end{itemize}

\section{Related Work} 
Image dehazing is to convert a haze image to a dehazed image. There are primarily two types of image dehazing methods, prior-based methods and learned-based methods. Recently, large convolutional kernel have become popular due to its efficiency and usefulness. It has large receptive fields and long-distance modeling capabilities that the vanilla convolution kernel does not have.

{\bfseries Prior-based image dehazing}: Early image dehazing methods were mainly based on prior knowledge, and the dehazing rules were found by statistical analysis of haze and haze-free image pairs. DCP \cite{he2010single} proposes that the minimum value of the image channel in the local haze area always approaches 0, and estimates $t(x)$ and $A$. Color attenuation prior samples more than 500 images, the author obtained a linear formula for estimating $d(x)$. Rank-one Prior \cite{rank-one} proposes that $t(x)$ is close to a rank-1 matrix and intensity projection strategy to estimate $t(x)$. The time complexity of prior-based methods is often very low, and the results of image restoration are excellent when the prior conditions are met. However, the results of image restoration will be distorted when the prior conditions are not met.

{\bfseries Learning-based image dehazing}: Due to the development of deep learning and the emergence of large image dehazing datasets, image dehazing methods based on deep learning have made great progress. DehazeNet \cite{Dehazenet} and MSCNN \cite{MSCNN} are early image dehazing networks that used neural networks to estimate $t(x)$ and prior-based methods to estimate $A$. DCPDN \cite{DCPDN} uses neural networks to estimate $A$ and $t(x)$ respectively. GridDehazeNet \cite{Griddehazenet} uses a grid-like neural network to obtain the multi-scale features of the image and estimate the haze-free image directly. It first points out that the estimation of haze-free images directly is better than the estimation of atmospheric scattering parameters. FFA-Net \cite{qin2020ffa} improved the effect of image dehazing by using a lot of channel Attention and pixel attention. AECR-Net\cite{AECR-net} improved the effect of image dehazing by introducing contrast learning. PMNet\cite{PMNet} using a novel Separable Hybrid Attention (SHA) module and a density map to effectively capture the unevenly distributed degeneration at the feature level. UDN\cite{UDN} uses an Uncertainty Estimation Block (UEB) to predict uncertainties and an Uncertainty-aware Feature Modulation (UFM) block to enhance learned features. With the excellent performance of transformers in image high-level tasks, many papers have recently used transformers in image dehazing tasks. DeHamer \cite{Dehamer} mixed transformer and CNN introduced the haze density into the transformer as absolute position embedding for the first time. Dehazeformer \cite{song2022vision} referred to the Swin transformer and modifies the key structure of the Swin transformer to make it more suitable for image dehazing. The transformer-based model has a large number of parameters, high latency, and is difficult to train. Therefore, we focus on the CNN-based method and use large dilated convolution to obtain the large receptive field and long-distance modeling capability possessed by the transformer. 

{\bfseries Large convolutional kernels}: RepLKNet \cite{largeKernel} proposes that using a few large convolutional kernels instead of a stack of small kernels could be a more powerful paradigm. It is a pure CNN architecture whose kernel size is $31\times31$ and points out large kernel CNNs have much larger effective receptive fields and higher shape bias rather than texture bias. RepLKNet \cite{largeKernel} achieves comparable or superior results than the Swin Transformer on ImageNet and a few typical downstream tasks, with lower latency. The Visual Attention Network \cite{VAN} proposes that a large kernel convolution can be divided into three components: a spatial local convolution (depth-wise convolution), a spatial long-range convolution (depth-wise dilation convolution), and a channel convolution (1×1 convolution) to overcome the huge amount of computational overhead and parameters. It surpasses similar-size vision transformers (ViTs) and convolutional neural networks (CNNs) in various vision tasks.     

\begin{figure*}
    \centering
    \includegraphics[width=17cm]{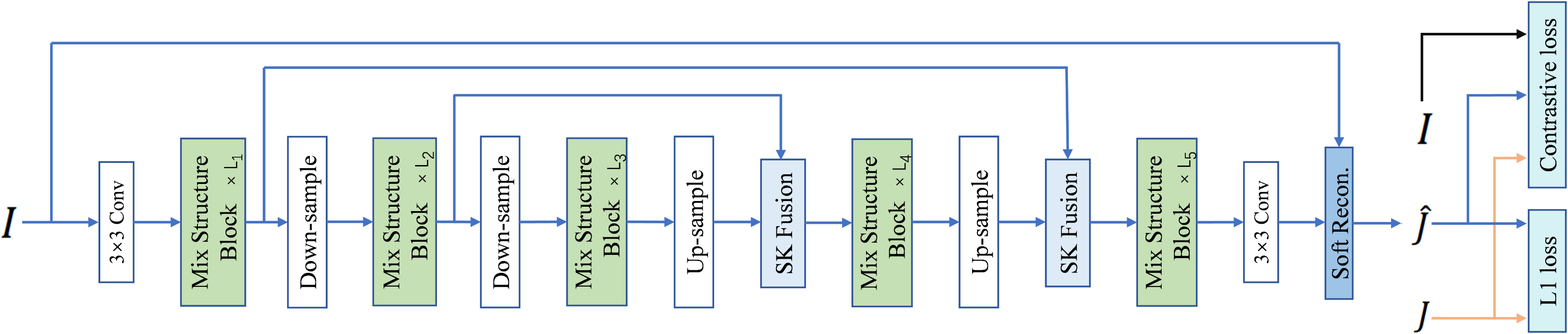}
     \caption{MixDehazeNet is a 5-stage U-net embedded in Mix Structure Block. Down-sample is 3$\times$3 Convolution with stride = 2. Up-sample is Point-Wise Convolution and PixelShuffle. {$I$} is the hazy image , {$J$} is the corresponding clear image and $\widehat{J}$ is the corresponding dehazing image.}
    \label{Figure 2}
    \centering
    \includegraphics[width=18cm]{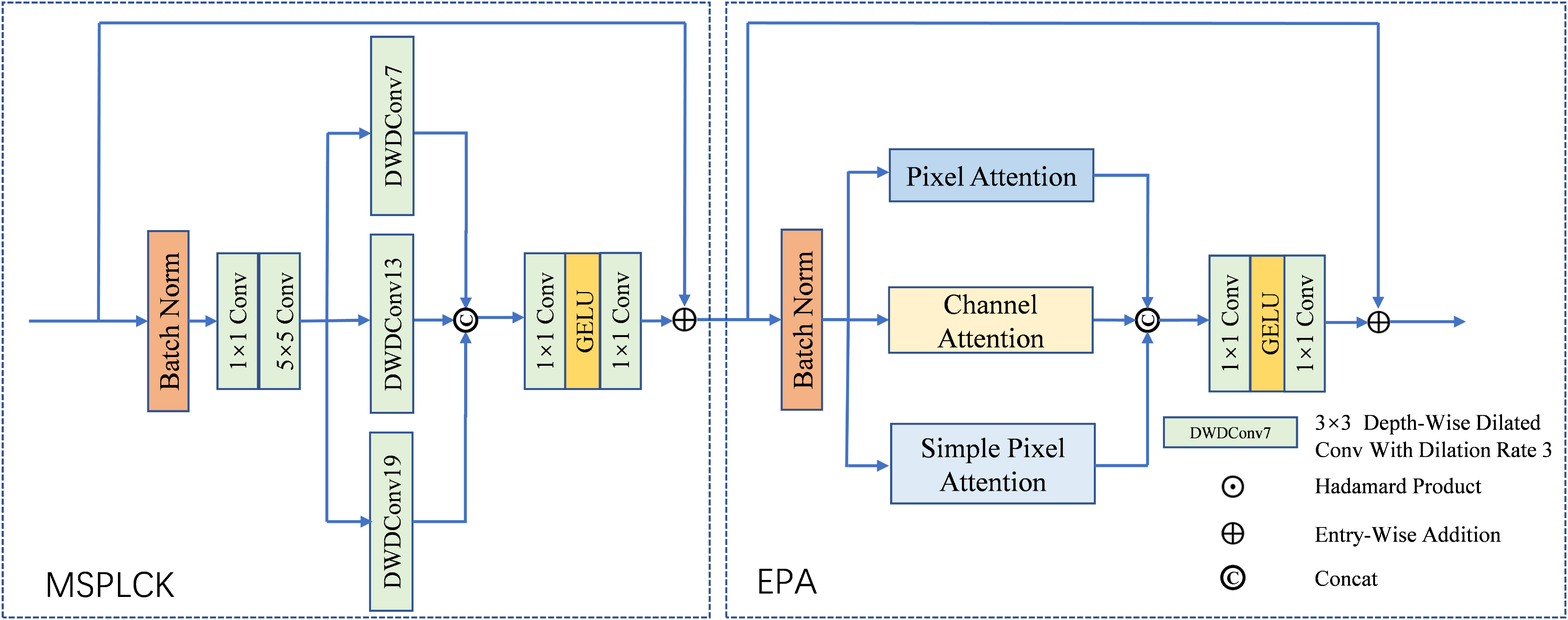}
     \caption{Mix Structure Block contains multi-scale parallel large convolution kernel module and enhanced parallel attention module.}
     \label{Figure 3}
\end{figure*}

\section{MixDehazeNet}
In this section, we mainly introduce our proposed dehazing network MixDehazeNet as shown in Figure \ref{Figure 2}. MixDehazeNet is a 5-stage U-net embedded in the Mix structure block which combines multi-scale parallel large convolution kernel and enhanced parallel attention. In addition, MixDehazeNet uses SK Fusion \cite{song2022vision} to fuse skip branches and main branches. And we use soft reconstruction \cite{song2022vision} instead of global residual at the end of the network, because soft reconstruction provides stronger haze removal constraints than global residual. 

\subsection{Multi-Scale Parallel Large Convolution Kernel}
Multi-scale Parallel Large Convolution Kernel module (MSPLCK) has both multi-scale characteristics and large receptive fields. First, let $x$ be the original feature map, we normalize it using BatchNorm via $\widehat{x}=BatchNorm(x)$. BatchNorm can accelerate network convergence, improve generalization ability and prevent overfitting.  
\begin{flalign}
\begin{aligned}
x1=&PWConv(\widehat{x}),\\
x2=&Conv(x1),\\
x3=&Concat(DWDConv19(x2),\\&DWDConv13(x2),\\&DWDConv7(x2))
\end{aligned}
\end{flalign}
Here, PWConv means point-wise convolution. Conv means convolution which kernel size = 5. DWDConv19 means which dilated convolution kernel size = 19 and it is $7\times7$ depth-wise dilated convolution with dilation rate 3, DWDConv13 means which dilated convolution kernel size = 13 and it is $5\times5$ depth-wise dilated convolution with dilation rate 3, DWDConv7 means which dilated convolution kernel size = 7 and it is $3\times3$ depth-wise dilated convolution with dilation rate 3. Finally, Concat means concatenate features in the channel dimension.

Three parallel dilated convolution with various kernel sizes can extract multi-scale features. And the large and medium dilated convolutions has long-distance modeling and large receptive fields like self-attention in transformer, they can concentrate on large haze areas. The small dilated convolution can concentrate on small haze areas and restore texture details. We concatenate multi-scale information from the channel dimension, and the feature dimension of $x3$ becomes three times of $x$. 
\begin{flalign}
\begin{aligned}
y=&x+PWConv(GELU(PWConv(x3)))
\end{aligned}
\end{flalign}

Then we feed $x3$ into a multi-layer perceptron that converts the feature dimension of $x3$ to be the same as $x$. The multi-layer perceptron contains two point-wise convolutions and uses GELU as an activate function. Finally, the output of the multi-layer perceptron is summed with the identity shortcut $x$. We believe that the multi-layer perceptron can not only combine three different types of features but also play a role in fitting the dehazing features.

\begin{figure}[t]
\centering
\subfigure[SPA]
{
    \begin{minipage}[b]{.4\linewidth}
        \centering
        \includegraphics[scale=0.07,angle=270]{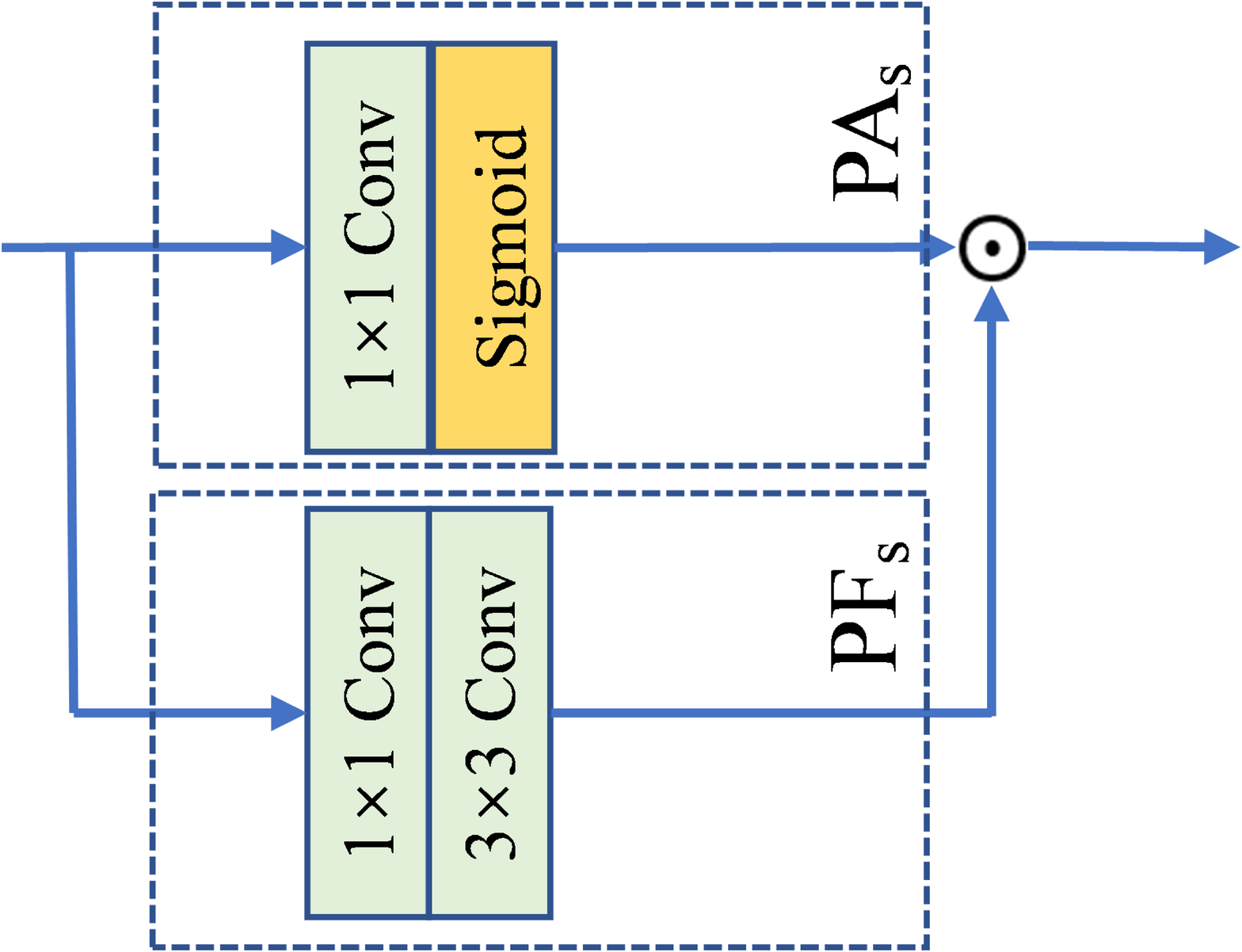}
        \label{Figure 4a}
    \end{minipage}
}
\subfigure[CA]
{
    \begin{minipage}[b]{.25\linewidth}
        \centering
        \includegraphics[scale=0.07,angle=270]{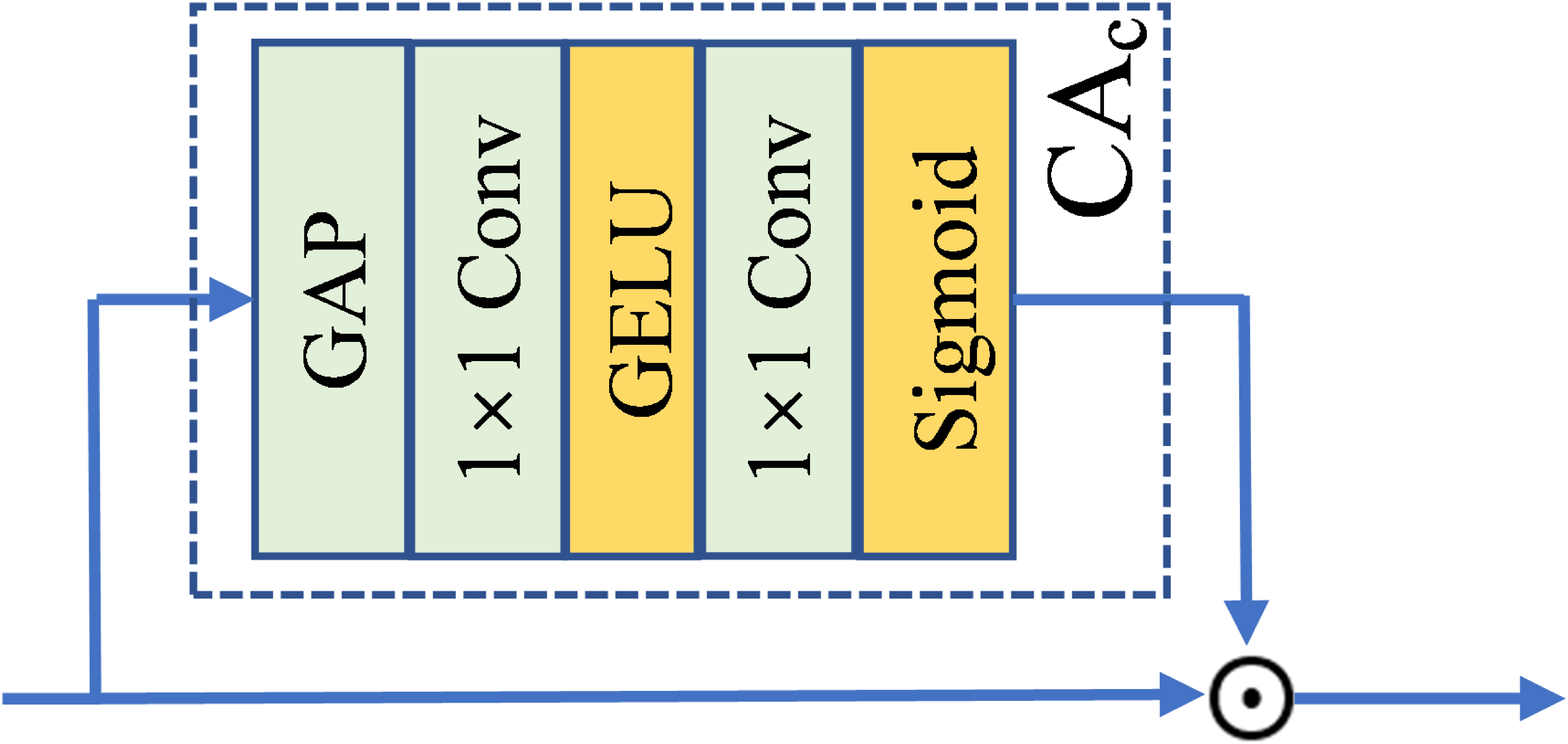}
        \label{Figure 4b}
    \end{minipage}
}
\subfigure[PA]
{
    \begin{minipage}[b]{.25\linewidth}
        \centering
        \includegraphics[scale=0.07,angle=270]{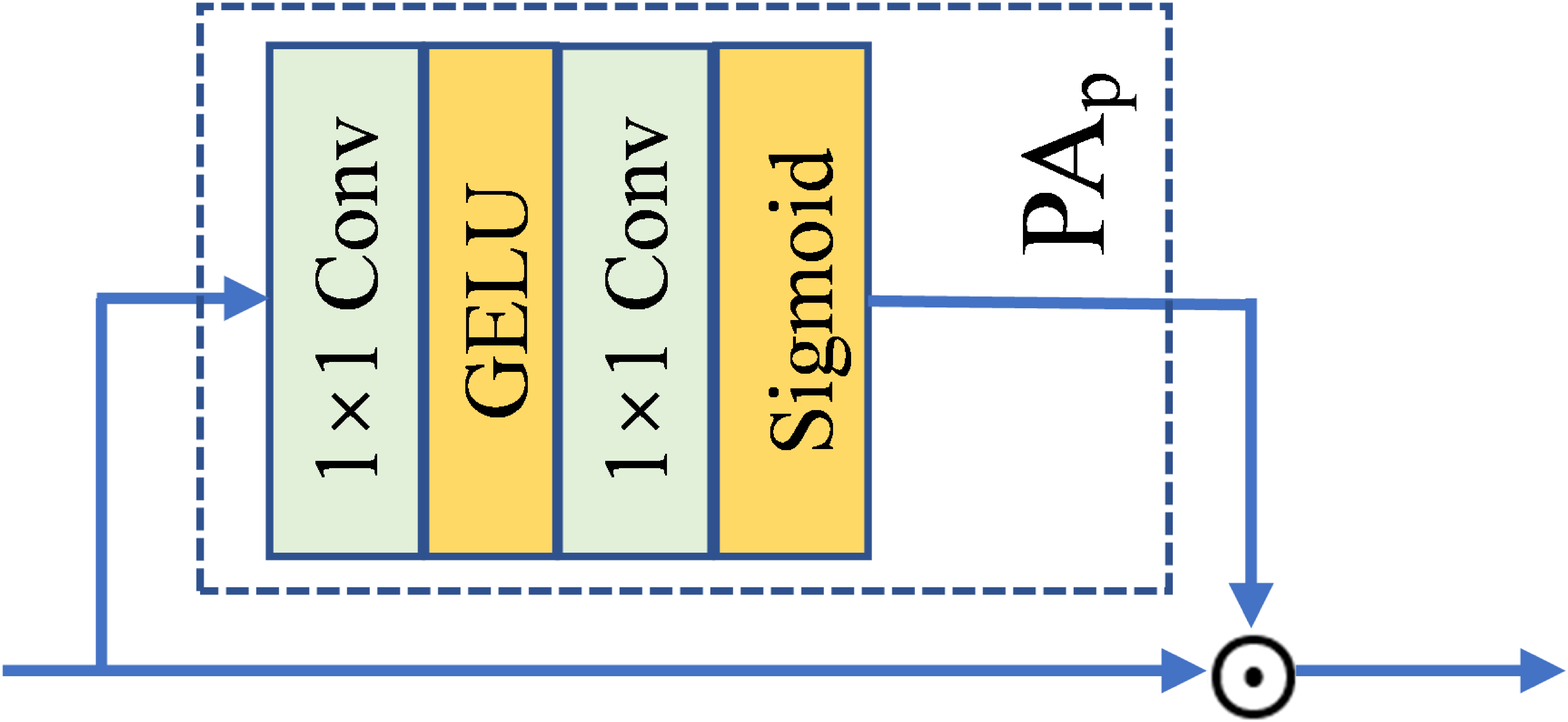}
        \label{Figure 4c}
    \end{minipage}
}
\caption{The schematic diagrams of Simple Pixel Attention (SPA), Channel Attention (CA) and Pixel Attention (PA). GAP is the global average pooling. }
\end{figure}
\subsection{Enhanced Parallel Attention}
Enhanced parallel attention module (EPA) mixed different types of attention mechanisms. It contains a simple pixel attention, a channel attention and a pixel attention. Let $x$ be the feature map, we normalize it using BatchNorm via $\widehat{x}=BatchNorm(x)$.

Pixel attention can effectively extract location-dependent informative features, such as different haze distributions across the image.  The simple pixel attention module consists of two branches: $PF_{s}$ and $PA_{s}$ as shown in Figure \ref{Figure 4a}. The $PF_{s}$ is a feature extraction branch. The $PA_{s}$ is a pixel gate branch. We use $PA_{s}$ as a pixel gating signal for $PF_{s}$. %\textcolor{red}{marked $PA_{s}$, $PA_{p}$ and $CA_{c}$ in the figure 3 and 4}
\begin{flalign}
\begin{aligned}
PF_{s}&=Conv(PWConv(\widehat{x})),\\
PA_{s}&=Sigmoid(PWConv(\widehat{x})),\\
F_{s}&=PF_{s} \otimes PA_{s}.\\
\end{aligned}
\end{flalign}
PWConv denotes point-wise convolution and Conv denotes convolution which kernel size = 3. The Pixel attention contains $PA_{p}$ branch, which can extract global pixel gating features. It is shown in Figure \ref{Figure 4c}.

\begin{flalign}
\begin{aligned}
PA_{p}=&Sigmoid(PWConv(GELU\\&(PWConv(\widehat{x})))),\\
F_{p}=&\widehat{x} \otimes PA_{p}.\\
\end{aligned}
\end{flalign}
Here we use PWConv-GELU-PWConv which can fit features. Sigmoid function is to extract global pixel gating features. And then use $PA_{p}$ as a global pixel gating signal for $\widehat{x}$. 

%Channel attention can effectively extract global information to change the channel dimension of feature. Standard channel attention contains $CA_{c}$ branch which can extract global channel gating features. 
Channel attention can efficiently extract global information and alter the channel dimension of a feature. Channel attention has a $CA_c$ branch that can extract features for the entire channel. It is shown in Figure \ref{Figure 4b}.
\begin{flalign}
\begin{aligned}
CA_{c}=&Sigmoid(PWConv(GELU\\&(PWConv(GAP(\widehat{x}))))),\\
F_{c}=&\widehat{x} \otimes CA_{c}.\\
\end{aligned}
\end{flalign}
We use the global average pooling (GAP), PWConv-GELU-PWConv, and Sigmoid function to extract global channel gating features. And then we use $CA_{c}$ as a global channel gating signal for $\widehat{x}$.
\begin{flalign}
\begin{aligned}
F=&Concat(F_{s},F_{c},F_{p}),\\
y=&x+PWConv(GELU(PWConv(F)))
\end{aligned}
\end{flalign}
We concatenate the three different attention gating results along the channel dimension. And then apply an MLP with PWConv-GELU-PWConv to reduce the concatenate feature channel dimension to the same dimension as the input $x$. Finally, the output of MLP is summed with the identity shortcut $\widehat{x}$.

Enhanced parallel attention module is more suitable for image dehazing. Atmospheric light $A$ is a shared global variable, while $t(x)$ is a location-dependent local variable. Channel attention can better extract shared global information and encode $A$. Pixel attention can better extract location-dependent information and encode $t(x)$. %We believe that when designing the attention mechanism, the global optimization can be achieved by extracting the share global information and location-dependent information of the original features at the same time. 
We think that by simultaneously extracting the location-dependent and share global information from the original features, the global optimization of the attention mechanism may be realized.
%Moreover, if two different types of attention mechanisms are connected in series, channel attention changes the original feature by extracting global information, and then extracting the changed feature's location-dependent information through pixel attention is not the global optimal situation. 
Nevertheless, when two distinct attention mechanisms are used in series, the global optimal condition is not achieved when channel attention modifies the original feature by extracting global information, and then pixel attention extracts the location-dependent information of the modified feature.
%We first parallel three different attention modules, so that the enhanced attention module can simultaneously extract the shared global variable and the location dependent local variable of the original feature. 
To enable the enhanced attention module to extract both the shared global variable and the location-dependent local variable of the original feature simultaneously, we parallel three different attention modules.
%We think that this method may better remove haze features based on encoding $A$ and $t(x)$ of the original feature in parallel, and then concatenate the three different types of attention results to get the combined feature $F$, then the merged feature $F$ is fused through an multi-layer perceptron. 
Based on encoding $A$ and $t(x)$ of the original feature in parallel, concatenating the three separate attention outcomes to obtain the combined feature $F$, and then fusing the combined feature $F$ through a multi-layer perceptron, we believe that this parallel module may better remove hazy features.

\subsection{Mix structure block}
Figure \ref{Figure 3} shows Mix Structure Block, which is a transformer-style block that includes a multi-scale parallel large convolution kernel module and an enhanced parallel attention module. The multi-scale parallel large convolution kernel module is used to obtain the multi-scale characteristics of the image which the single transformer \cite{vit,swinTransformer} multi-head self-attention module does not have. The enhanced parallel attention module can efficiently deal with an uneven hazy distribution which a single transformer \cite{vit,swinTransformer} feed-forward module does not have. The multi-scale parallel large convolution kernel module can simultaneously capture large areas of haze and restore texture details. The enhanced parallel attention module that can extract shared global information and location-dependent local information of the original feature in parallel. The proposed MixDehazeNet which contains Mix Structure Block has made state-of-the-art achievements on multiple image dehazing datasets.

\subsection{Training Loss}
Given the image pair {$I,J$} where {$I$} is hazy image and {$J$} is the corresponding clear image, we let MixDehazeNet predict the dehazing image $\widehat{J}$.  And we use $L_{1}$ loss and contrastive loss to train our model, which can be formulated as:
\begin{flalign}
\begin{aligned}
min||J-\widehat{J}||_{1}+\beta\sum_{i=0}^n\omega_{i}\cdot\frac{D(R_{i}(J),R_{i}(\widehat{J}))}{D(R_{i}(I),R_{i}(\widehat{J}))}
\end{aligned}
\end{flalign}
where $R_{i}, i=1,2,\cdot\cdot\cdot,n$ extracts the $i$-th layer features from the fixed pre-trained model. $D(x,y)$ is the $L_{1}$ loss. $\omega_{i}$ is a weight coeffcient. $\beta$ is hyper parameter to balance $L_{1}$ loss and contrastive learning loss.

\begin{table*}[t]
\small
\setlength{\tabcolsep}{1.0mm}{
\centering
\fontsize{11}{19}\selectfont
\caption{\centering Quantitative comparison of various SOTA methods on three dehazing datasets.}
\label{Table 1}
\begin{tabular}{|c|c|c|c|c|c|c|c|c|c|}
\hline
\multirow{10}{*}{Text} &\multicolumn{2}{c|}{RESIDE-IN}&\multicolumn{2}{c|}{RESIDE-OUT} &\multicolumn{2}{c|}{RESIDE-6K}&\multicolumn{3}{c|}{OVERHEAD}\cr\hline
& PSNR & SSIM & PSNR & SSIM & PSNR&SSIM&Param(M)& MACs(G)& Latency(ms) \cr
DCP\cite{he2010single}(TPAMI'10) & 16.62 &0.818 & 19.13 & 0.815 & 17.88 & 0.816 &-&-&-\cr
DehazeNet\cite{Dehazenet}(TIP'16) & 19.82& 0.821 & 24.75& 0.927 & 21.02& 0.870 &0.009&0.581&0.919\cr
MSCNN\cite{MSCNN}(ECCV'16) & 19.84& 0.833 & 22.06& 0.908 & 20.31& 0.863 &0.008&0.525&0.619\cr
AOD-Net\cite{AOD-Net}(ICCV'17) & 20.51& 0.816 & 24.14& 0.920 & 20.27& 0.855 &0.002&0.115&0.390\cr
GFN\cite{GFN}(CVPR'18) & 22.30& 0.880 & 21.55& 0.844 & 23.52& 0.905 &0.499&14.94&3.849\cr
GCANet\cite{GCANet}(WACV'19) & 30.23& 0.980 & -& - & 25.09& 0.944 &0.702&18.41&3.695\cr
GirdDehazeNet\cite{Griddehazenet}(ICCV'19) & 32.16& 0.984 & 30.86& 0.982 & 25.86& 0.944 &0.956&21.49&9.905\cr
MSBDN\cite{MSBDN}(CVPR'20) & 33.67& 0.985 & 33.48& 0.982 & 28.56& 0.966 &31.35&41.54&13.25\cr
PFDN\cite{PFDN}(ECCV'20) & 32.68& 0.976 & -& - & 28.15 & 0.962 &11.27&50.46&4.809\cr
FFA-Net\cite{qin2020ffa}(AAAI'20) & 36.39& 0.989 & 33.57& 0.984 & 29.96 & 0.973 &4.456&287.8&55.91\cr
AECR-Net\cite{AECR-net}(CVPR'21) & 37.17& 0.990 & -& - & 28.52 & 0.964 &2.611&52.20&28.08\cr
PMNet\cite{PMNet}(ECCV'22) & 38.41& 0.990 & 34.74& 0.985 & -& - &18.90&81.13&27.16\cr
MAXIM\cite{MAXIM}(CVPR’22)& 38.11 & 0.991 & 34.19 &  0.985 & - & - &14.1&216&-\cr
Dehamer\cite{Dehamer}(CVPR’22)& 36.36 & 0.988 & 35.18 & 0.986 & - & - &132.45&48.93&-\cr
UDN\cite{UDN}(AAAI’22)& 38.62 & 0.991 & 34.93 & 0.987 & - & - &4.25&-&-\cr
\hline
% MixDehazeNet-T &38.64 & 0.994 & &  & &  &&\cr
MixDehazeNet-S & 39.47 & 0.995 & 35.09 & 0.985 &\textbf{30.18} &\textbf{0.973} &3.16&22.06&14.56\cr
MixDehazeNet-B & 40.90 & 0.996 & 35.67 & 0.985 &-&-&6.25&43.61&28.45\cr
MixDehazeNet-L & \textbf{42.62} &\textbf{0.997} & \textbf{36.50} & \textbf{0.986} &-&-&12.42&86.7&56.52\cr
\hline
\end{tabular}}
\end{table*}

% Compare image quality in RESIDE-IN

%-------------------------------------------------------------------------
\section{Experiments}

\subsection{Datasets}
We evaluated our method on the RESIDE \cite{dataset}, RESIDE-6K datasets. RESIDE\cite{dataset} is the most standard datasets of image dehazing. The RESIDE datasets contains of RESIDE-IN(ITS), RESIDE-OUT(OTS) and Synthetic Objective Testing task(SOTS). The RESIDE-6K dataset contains a mix of synthesis images of indoor and outdoor scenes from ITS and OTS.

%1) We use RESIDE-IN (13990 image pairs) to train the models and test them on the indoor set (500 image pairs) of the SOTS. We train MixDehazeNet on ITS for 500 epochs. 
1) We trained our models on RESIDE-IN, which contains 13,990 image pairs, and tested them on the indoor set (500 image pairs) of SOTS. MixDehazeNet was trained on ITS for 500 epochs.

%2) We use RESIDE-OUT (313950 image pairs) to train the models and test them on the outdoor set (500 image pairs) of the SOTS. We train MixDehazeNet on OTS for 40 epochs. 
2) We trained our models on RESIDE-OUT, which contains 313,950 image pairs, and tested them on the outdoor set (500 image pairs) of SOTS. MixDehazeNet was trained on OTS for 40 epochs.

%3) RESIDE-6K dataset contains 6000 image pairs which 3000 ITS image pairs and 3000 OTS image pairs are used for training and the remaining 1000 image pairs which mixes indoor and outdoor image pairs for testing. We use an experimental setup from DA \cite{DA}. We train MixDehazeNet on RESIDE-6k for 1000 epochs.
3) The RESIDE-6K dataset contains 6,000 image pairs which 3,000 ITS image pairs and 3,000 OTS image pairs are used for training, and the remaining 1,000 image pairs which mix indoor and outdoor image pairs are used for testing. We used an experimental setup from DA \cite{DA} and trained MixDehazeNet on RESIDE-6K for 1000 epochs.

\subsection{Implementation Details}
We used 4-card RTX-3090 to train our models. During training, images are randomly cropped to 256 × 256 patches. We provided three MixDehazeNet variants (-S, -B, -L for small, basic, and large, respectively). Table \ref{Table 2} lists the detailed configurations of the variants. We extracted the hidden features of 11th , 35th , 143rd , 152nd layers from the fixed pre-trained Resnet-152, and their corresponding coefficients $\omega_{i},i= 1,\cdot\cdot\cdot,4$ to $\frac{1}{16},\frac{1}{8},\frac{1}{4},1$. And we set the hyper parameter $\beta$ to 0.1. We used the AdamW optimizer to optimize our MixDehazeNet with exponential decay rates $\beta_{1}$ and $\beta_{2}$ equals to 0.9 and 0.999, respectively. We set the initial learning rate to $2 \times 10^{-4}$, which gradually decreases from the initial rate to $2 \times 10^{-6}$ with the cosine annealing strategy. 

\begin{figure*}[htbp]
\centering
\subfigure[Hazy]
{
    \begin{minipage}[b]{.15\linewidth}
        \centering
        \includegraphics[scale=0.13]{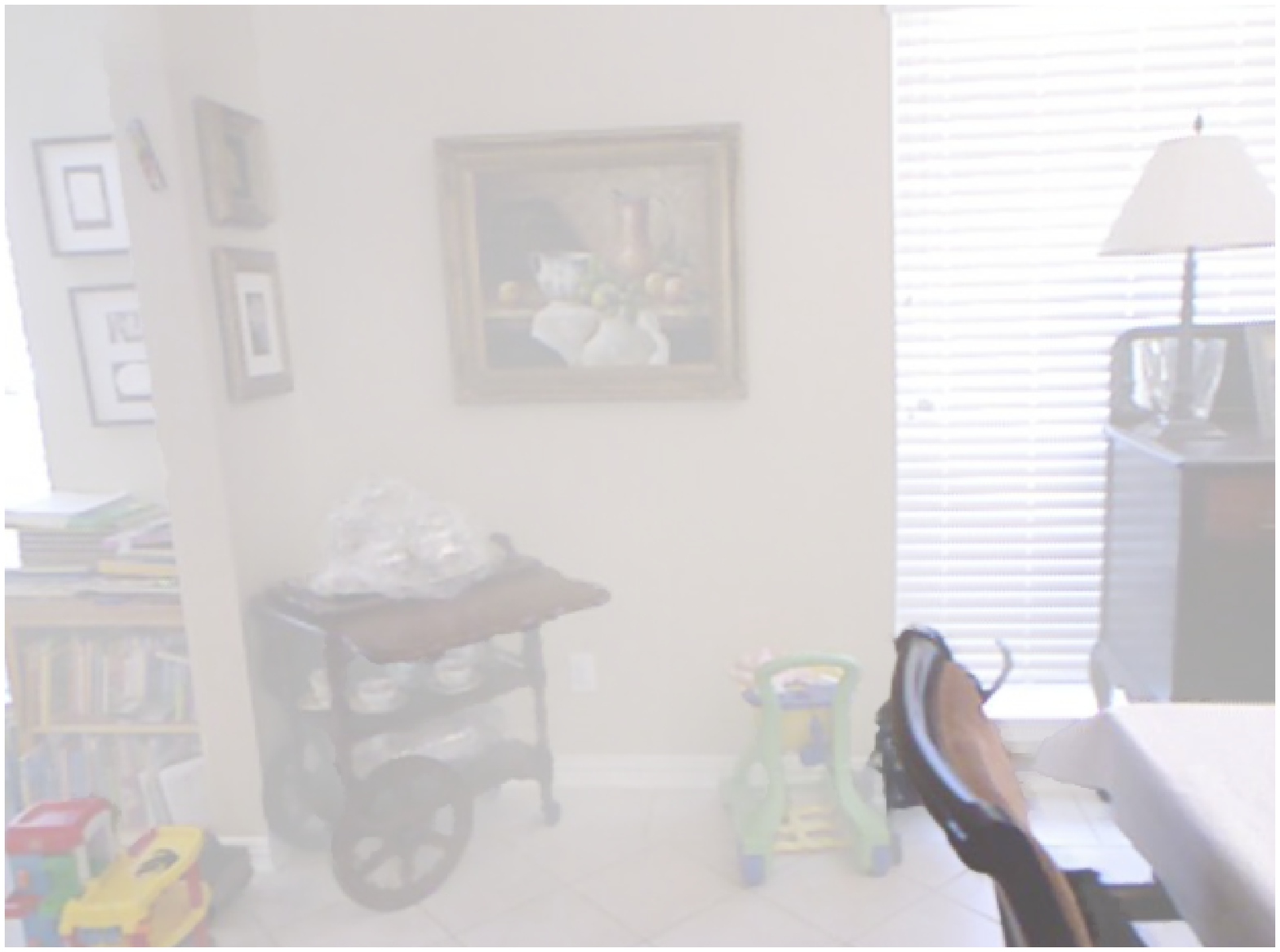} \\
        \includegraphics[scale=0.13]{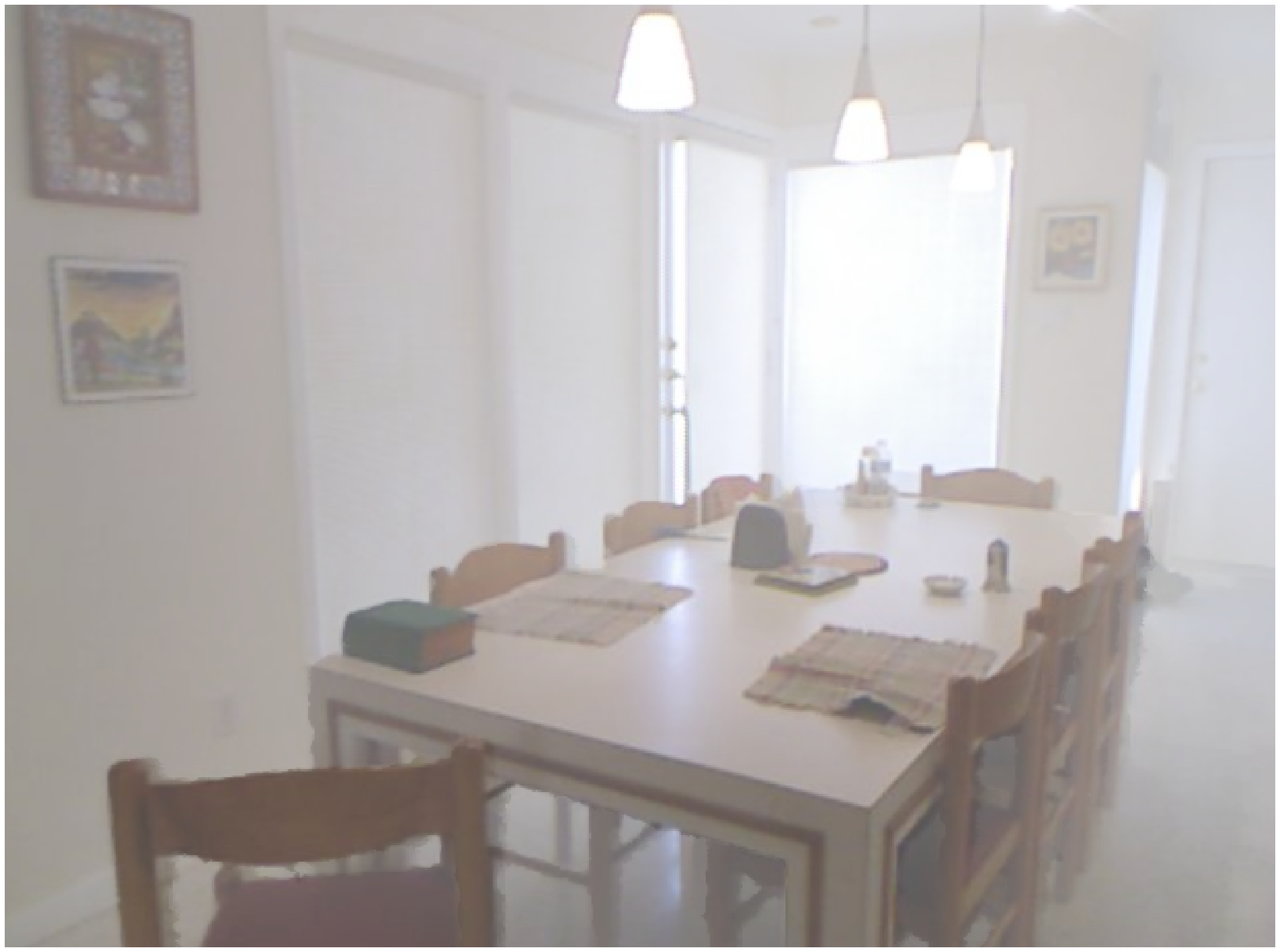}
    \end{minipage}
}
\subfigure[DCP\cite{DCPDN}]
{
    \begin{minipage}[b]{.15\linewidth}
        \centering
        \includegraphics[scale=0.13]{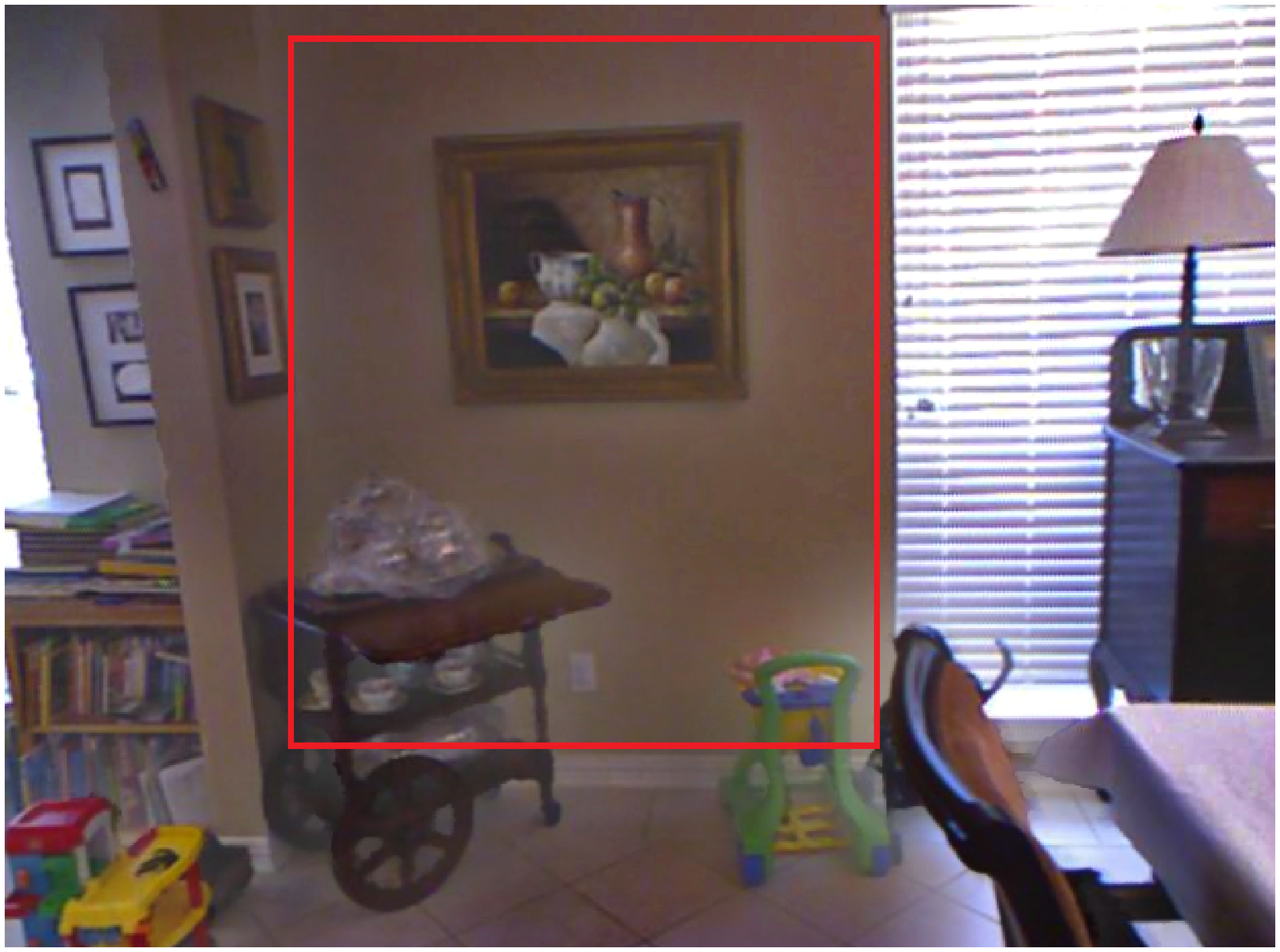} \\
        \includegraphics[scale=0.13]{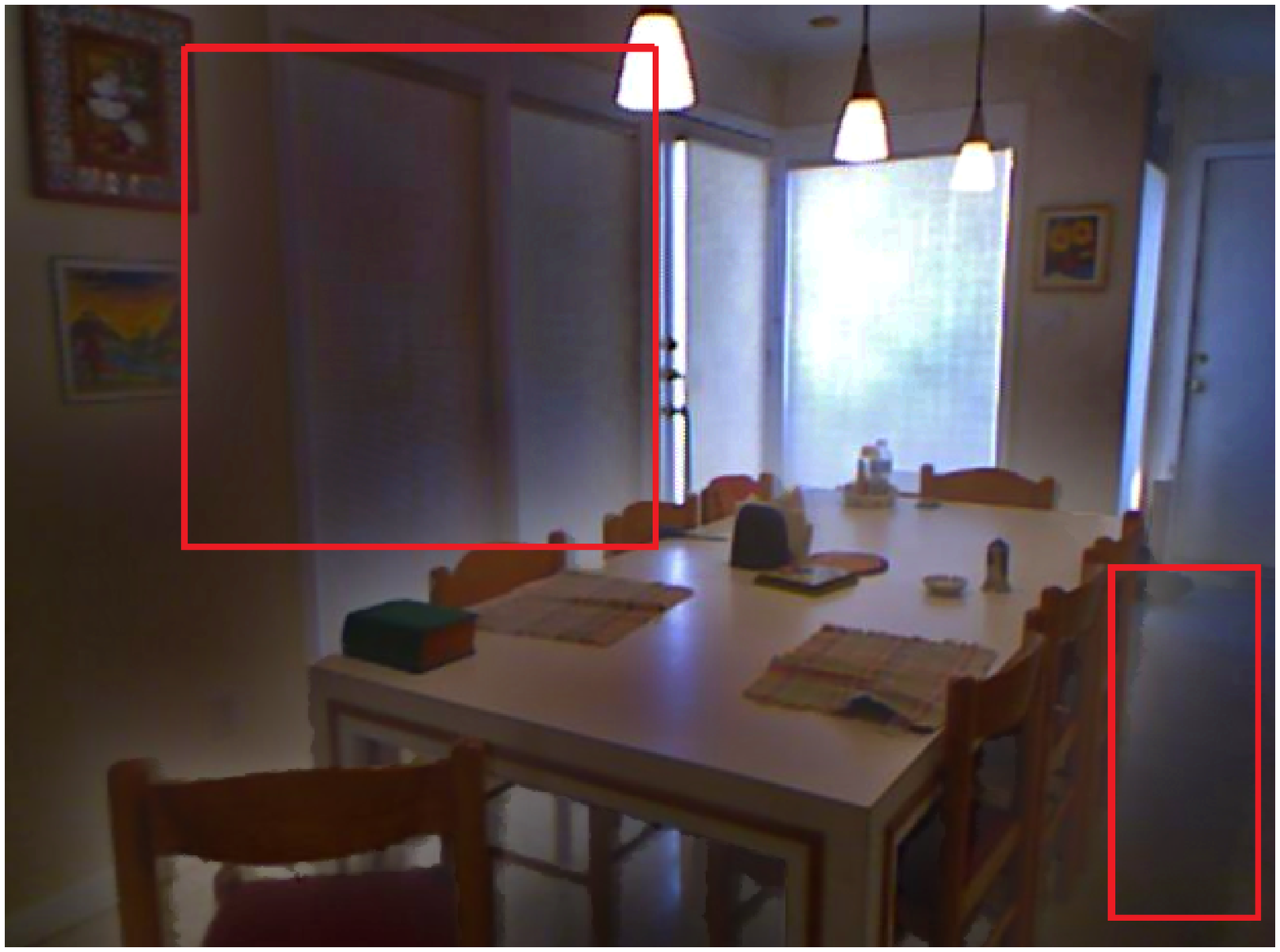}
    \end{minipage}
}
\subfigure[GridDehazeNet\cite{Griddehazenet}]
{
    \begin{minipage}[b]{.15\linewidth}
        \centering
        \includegraphics[scale=0.13]{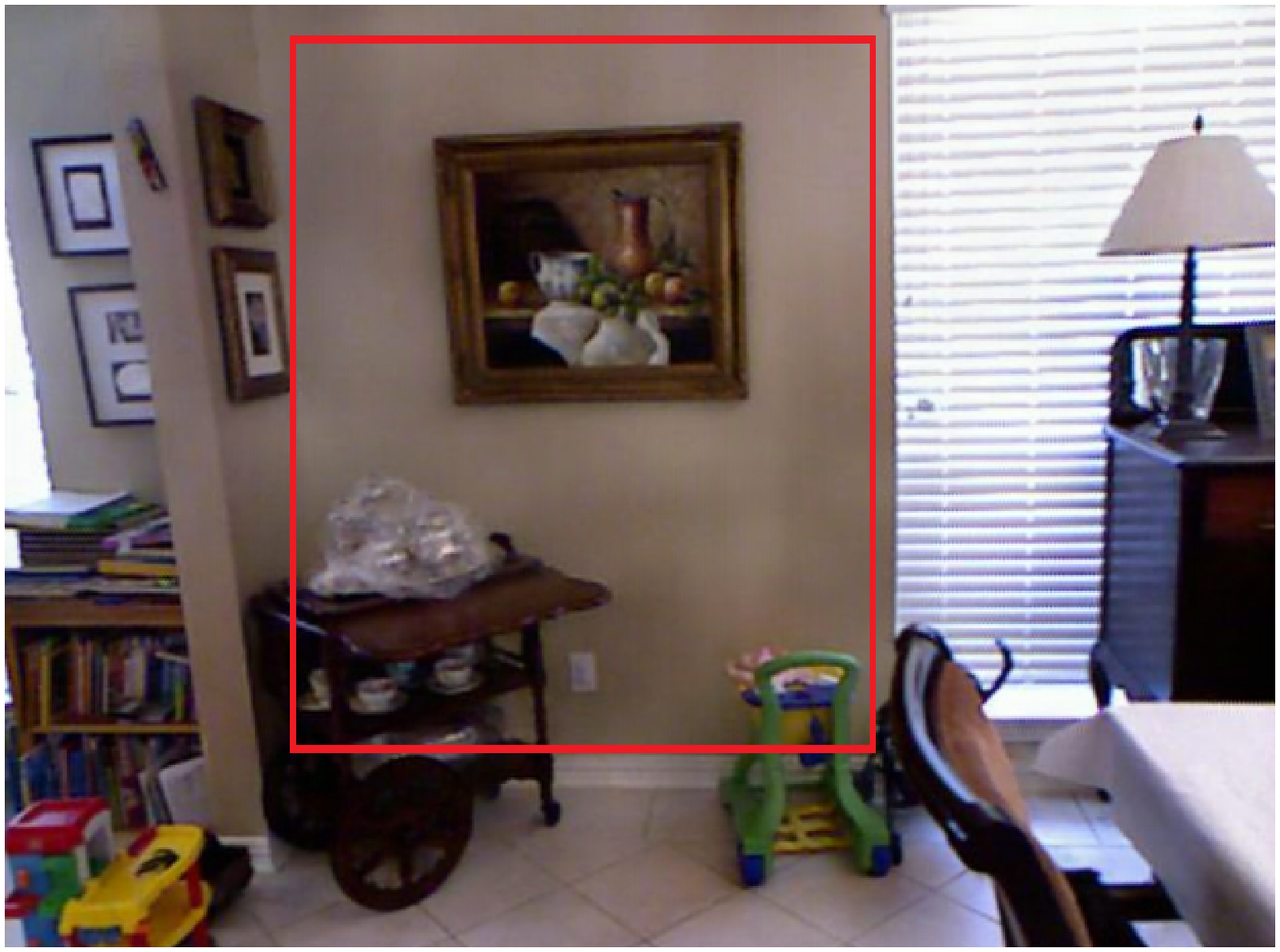} \\
        \includegraphics[scale=0.13]{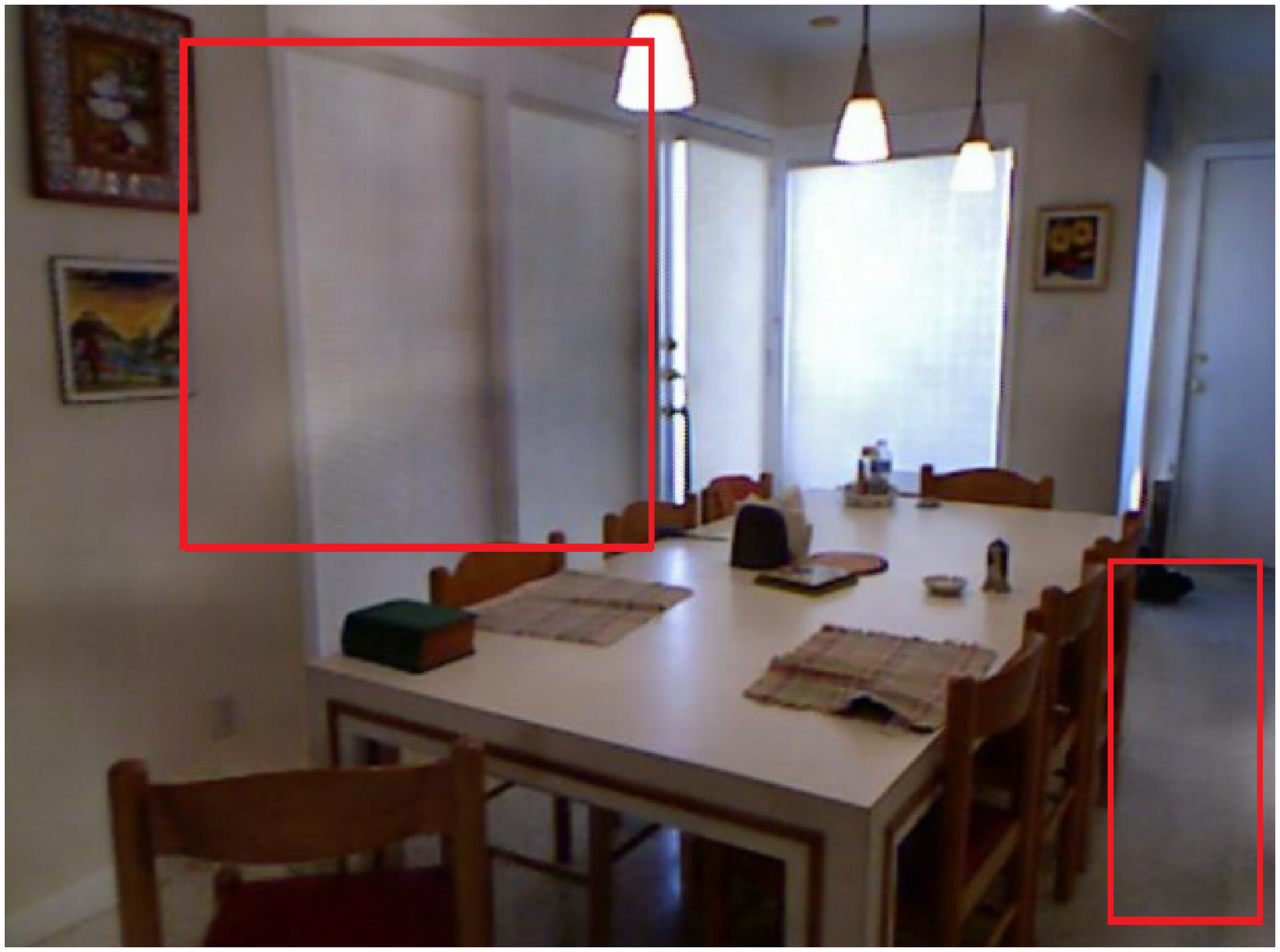}
    \end{minipage}
}
\subfigure[FFA-Net\cite{qin2020ffa}]
{
    \begin{minipage}[b]{.15\linewidth}
        \centering
        \includegraphics[scale=0.13]{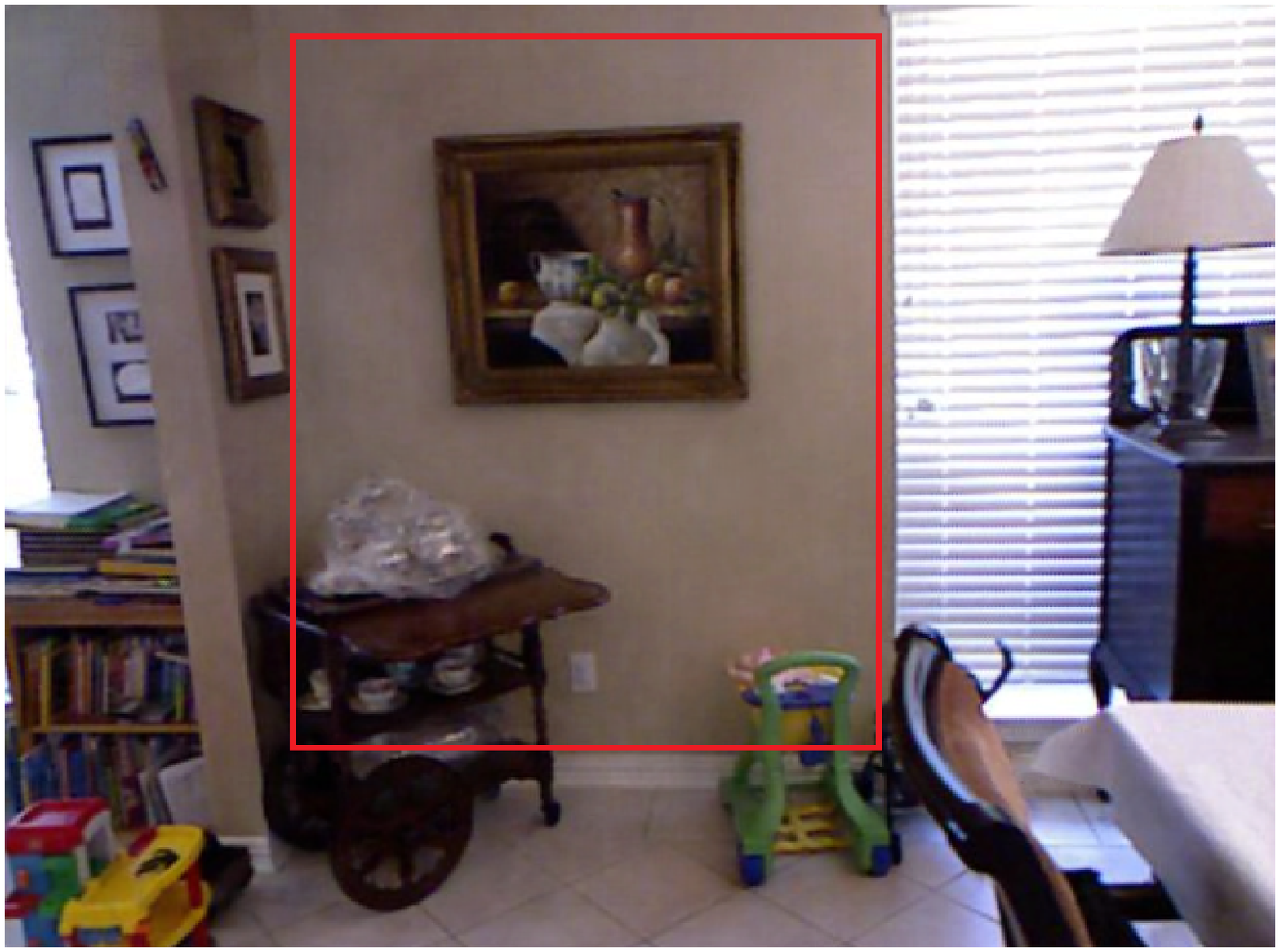} \\
        \includegraphics[scale=0.13]{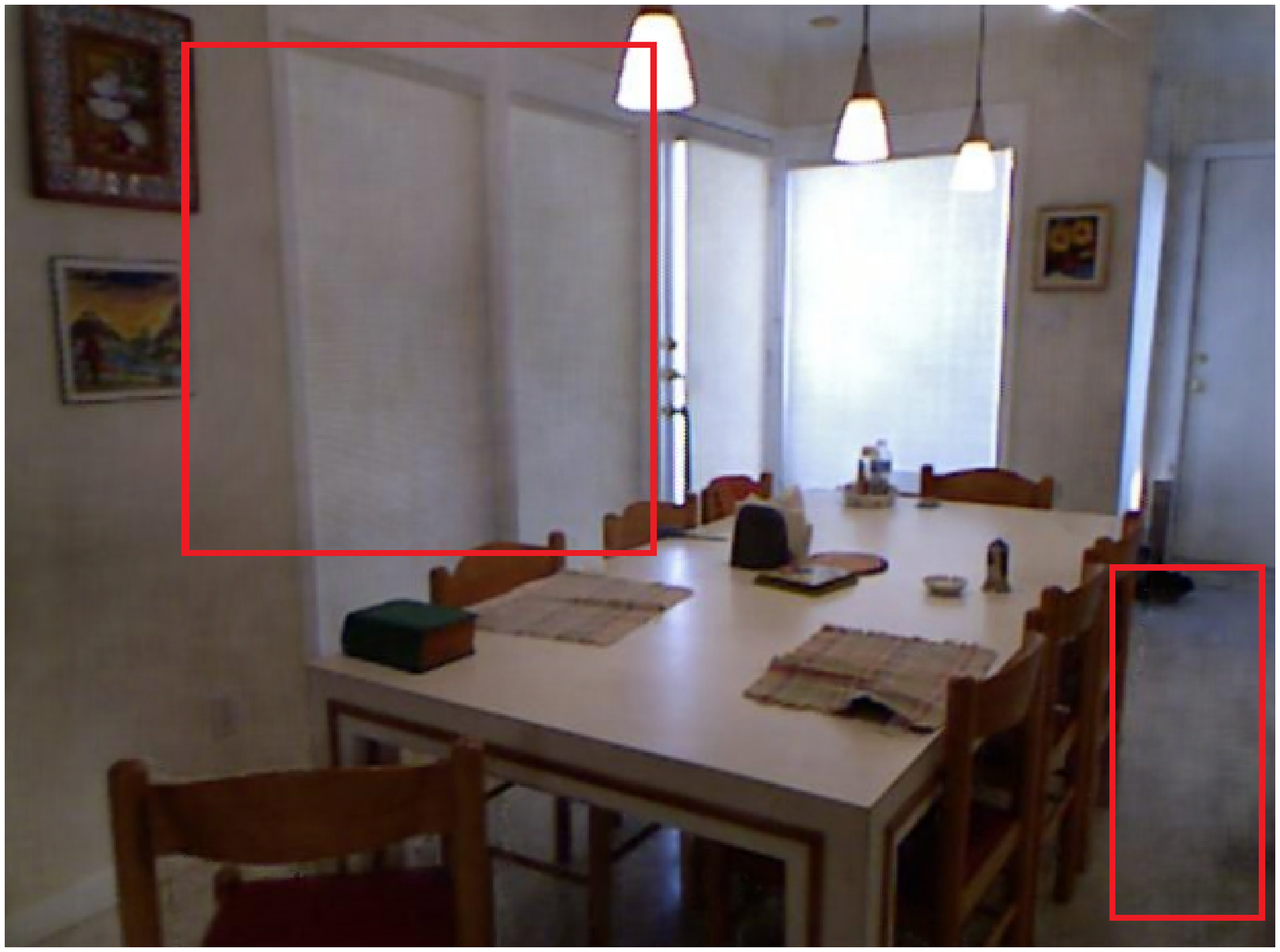}
    \end{minipage}
}
\subfigure[ours]
{
    \begin{minipage}[b]{.15\linewidth}
        \centering
        \includegraphics[scale=0.13]{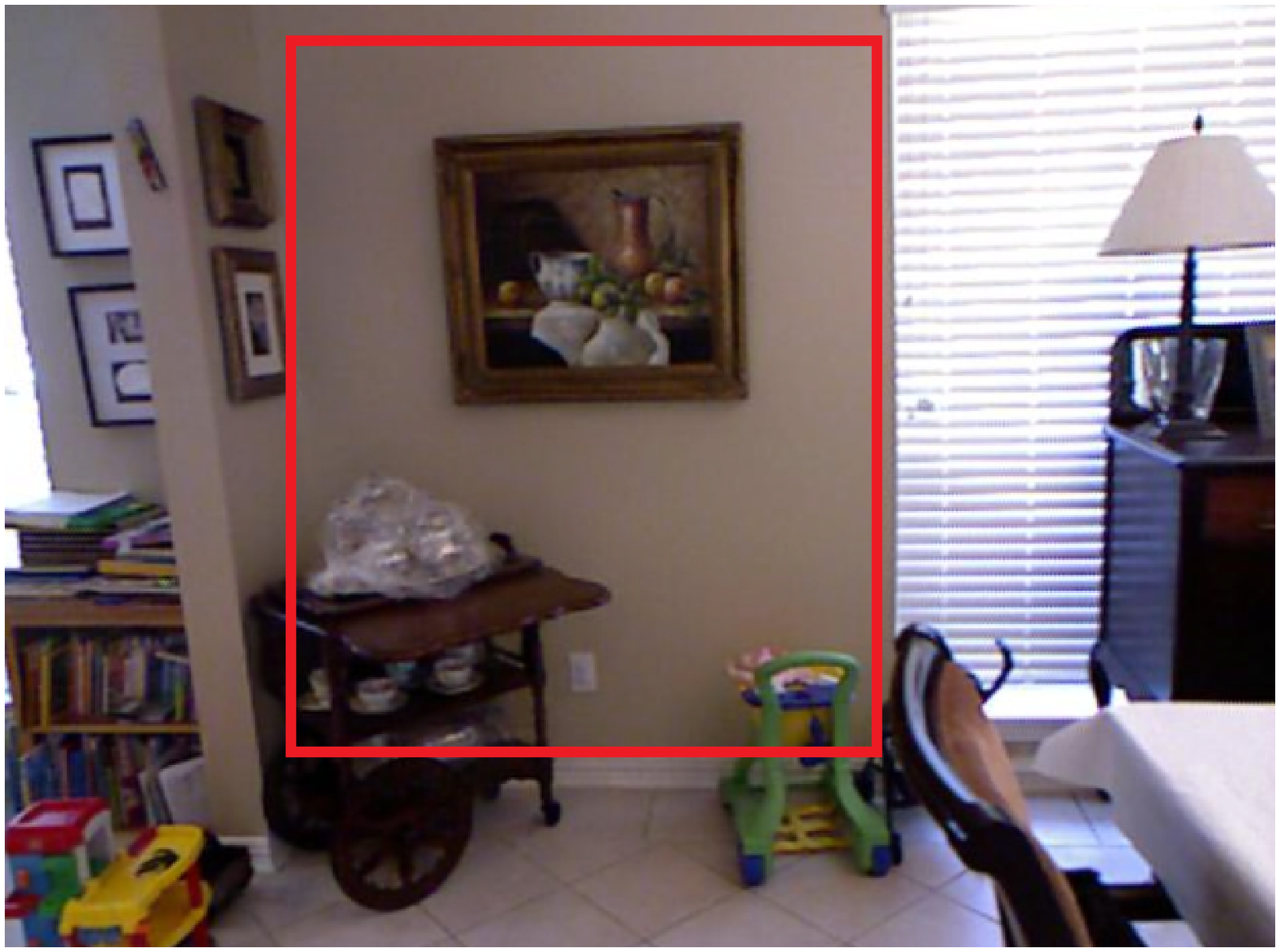} \\
        \includegraphics[scale=0.13]{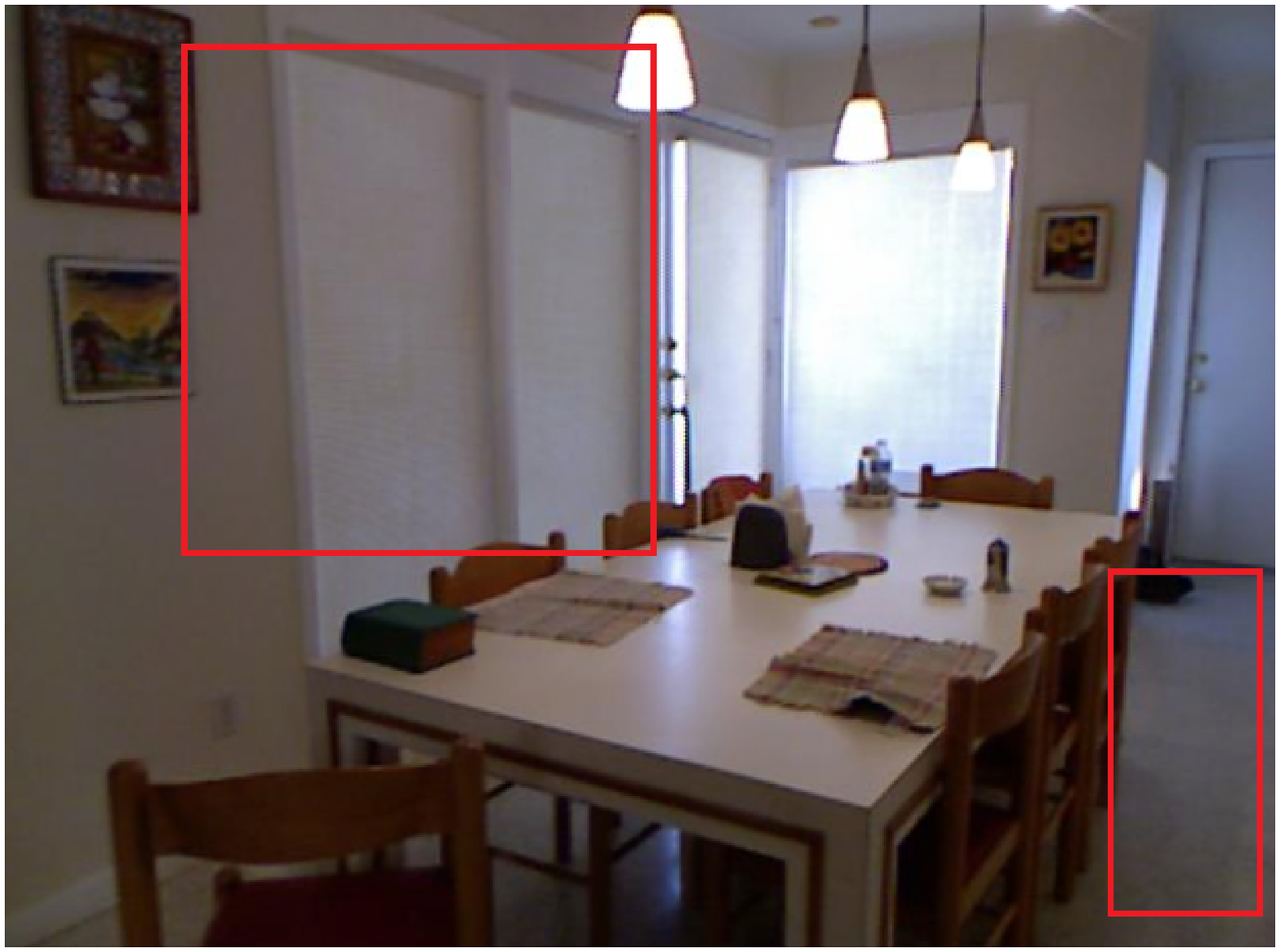}
    \end{minipage}
}
\subfigure[GT]
{
    \begin{minipage}[b]{.15\linewidth}
        \centering
        \includegraphics[scale=0.13]{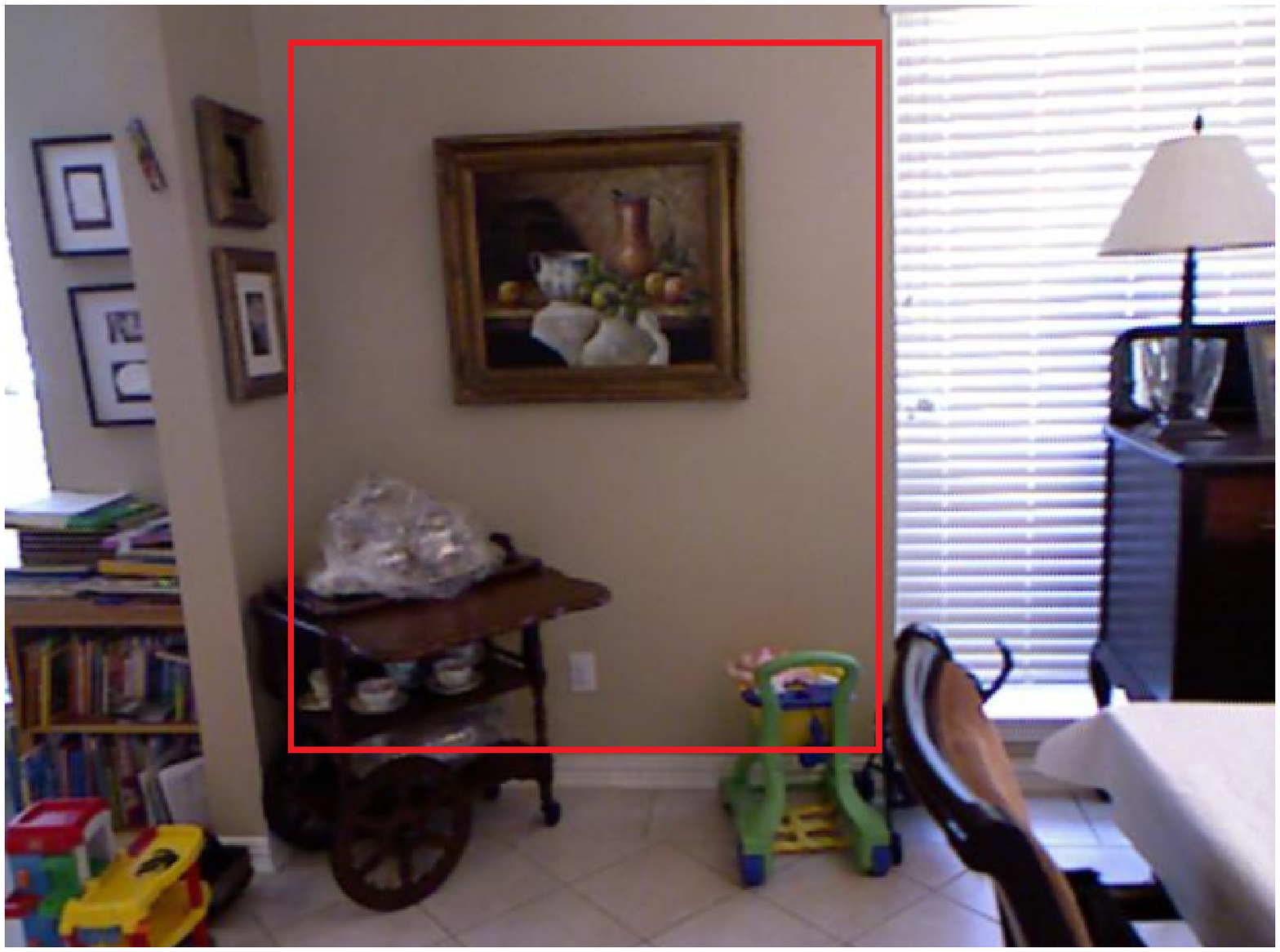} \\
        \includegraphics[scale=0.13]{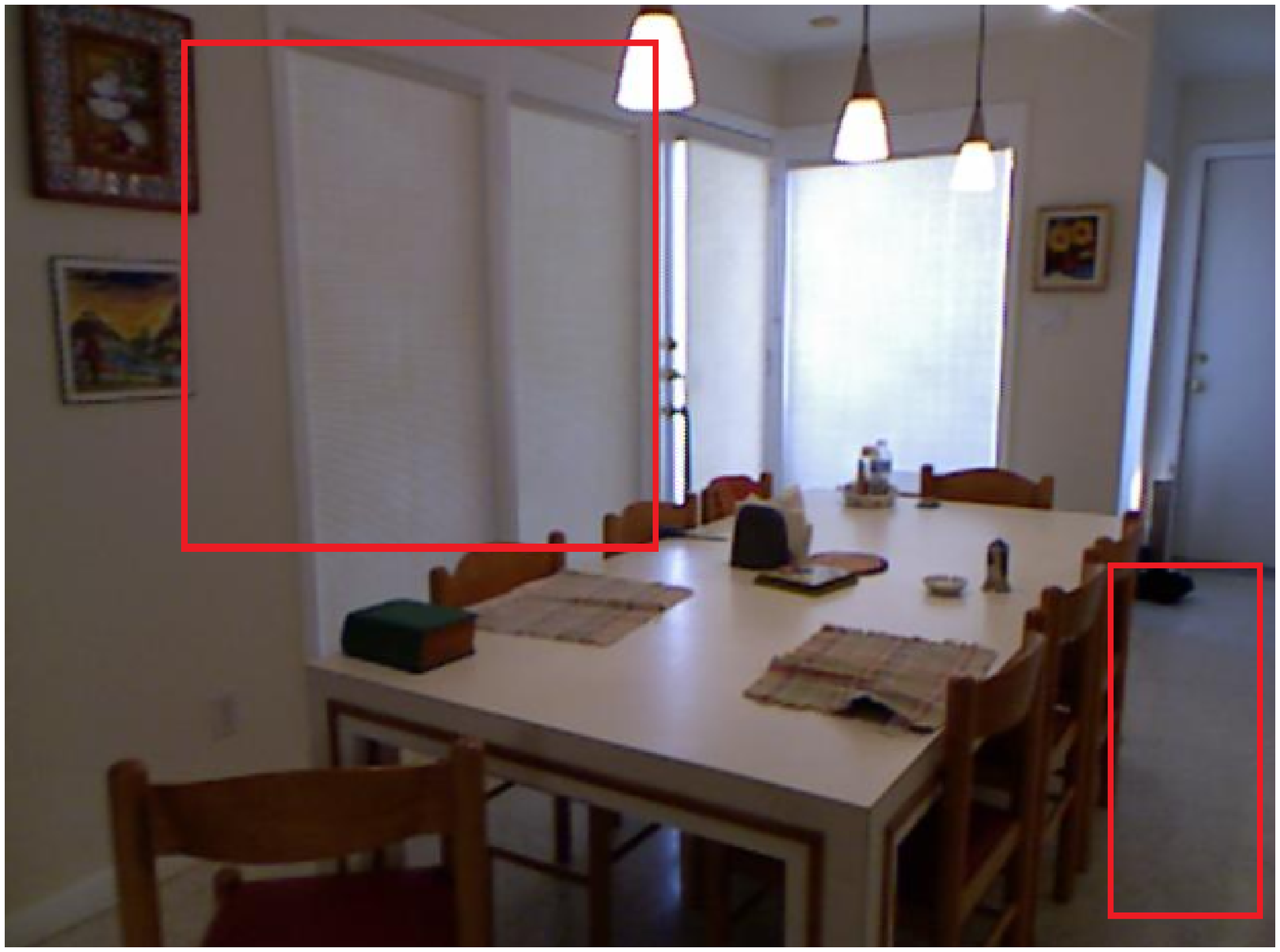}
    \end{minipage}
}
\caption{Qualitative comparisons on RESIDE-IN dataset. Zoom in for the best view.}
\label{Figure 5}
\end{figure*}

%compare image quality in RESIDE-OUT
\begin{figure*}[htbp]
\centering
\subfigure[Hazy]
{
    \begin{minipage}[b]{.15\linewidth}
        \centering
        \includegraphics[scale=0.147]{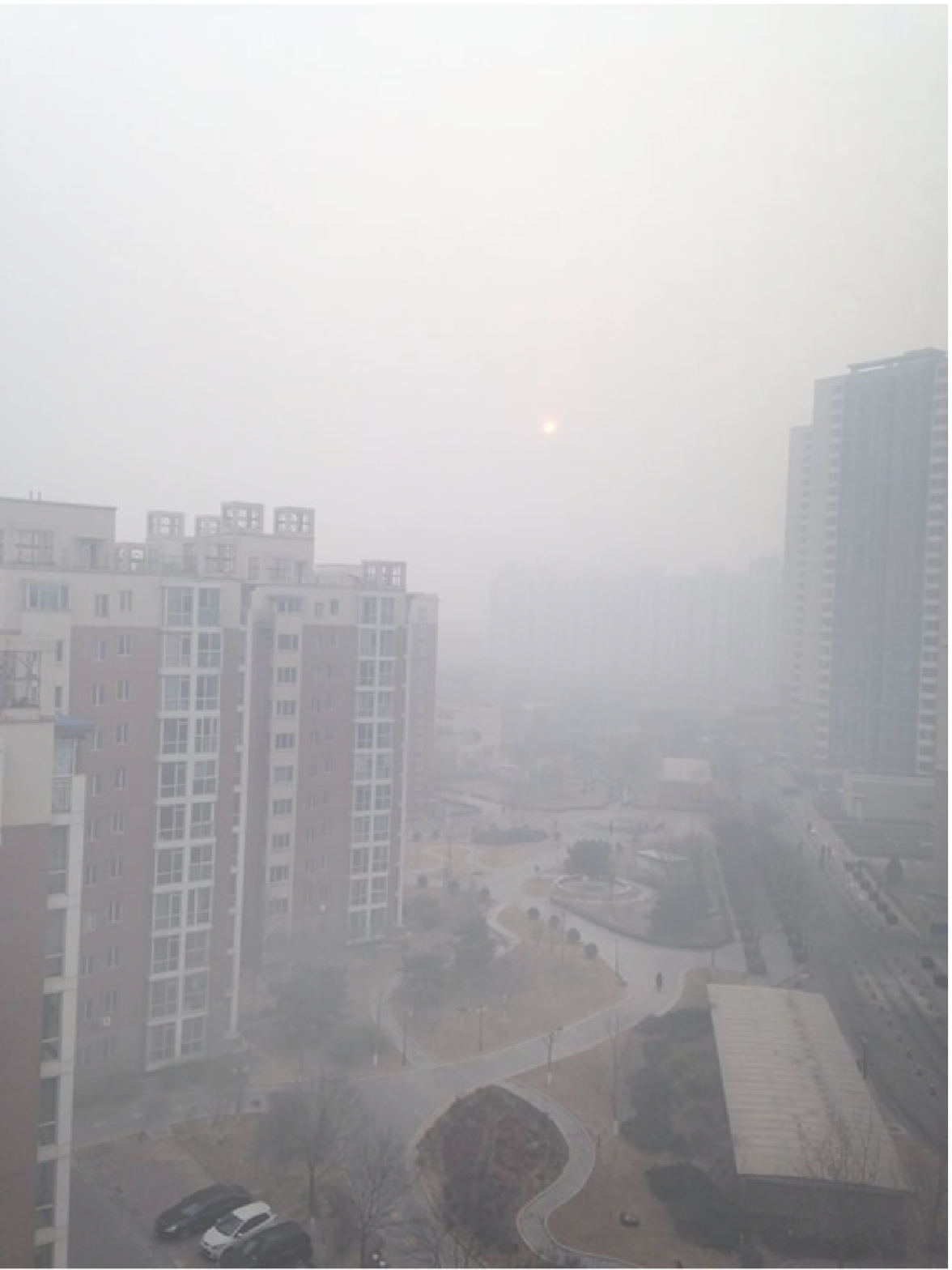}
    \end{minipage}
}
\subfigure[DCP\cite{DCPDN}]
{
    \begin{minipage}[b]{.15\linewidth}
        \centering
        \includegraphics[scale=0.147]{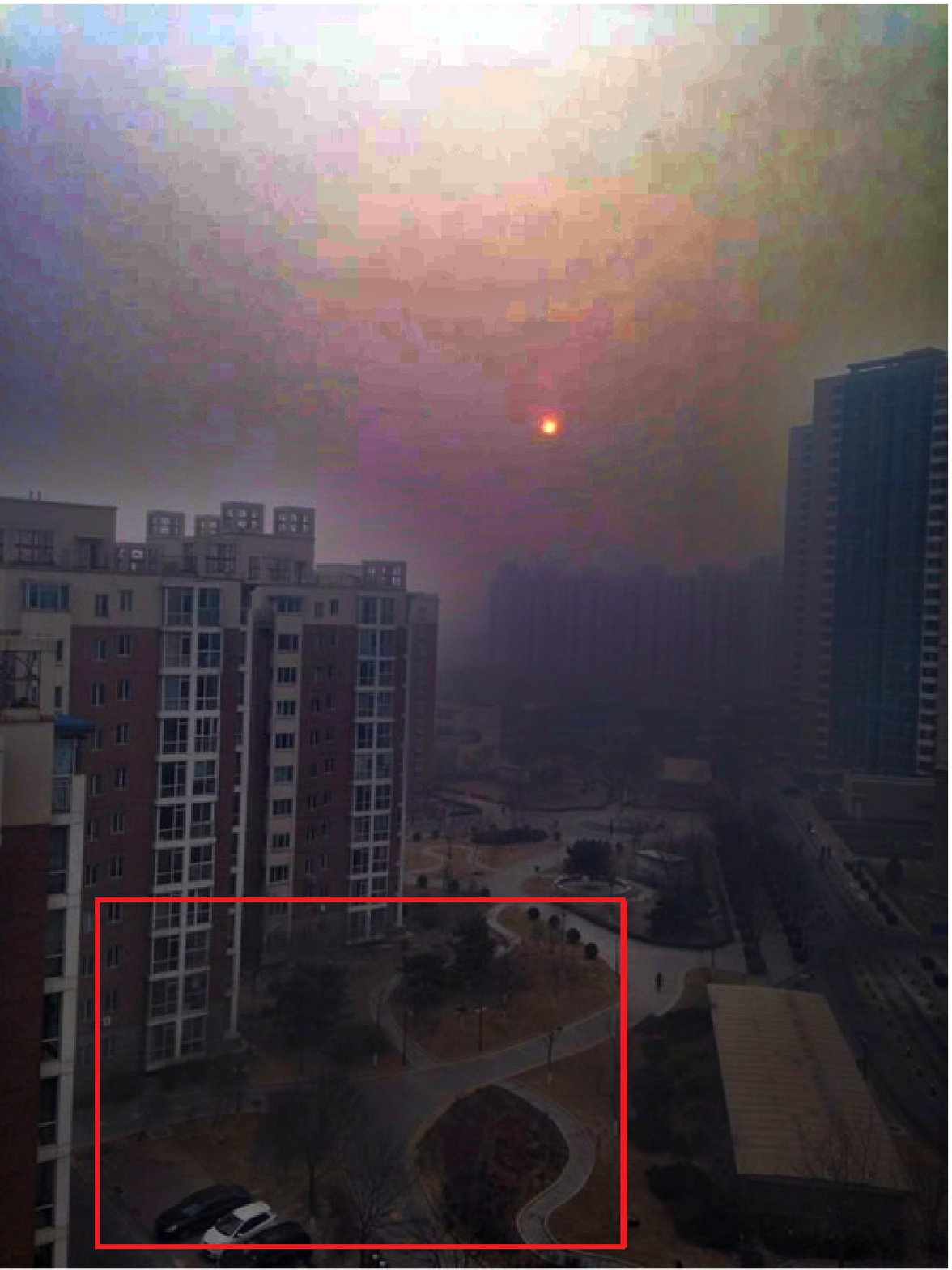}
    \end{minipage}
}
\subfigure[GridDehazeNet\cite{Griddehazenet}]
{
    \begin{minipage}[b]{.15\linewidth}
        \centering
        \includegraphics[scale=0.147]{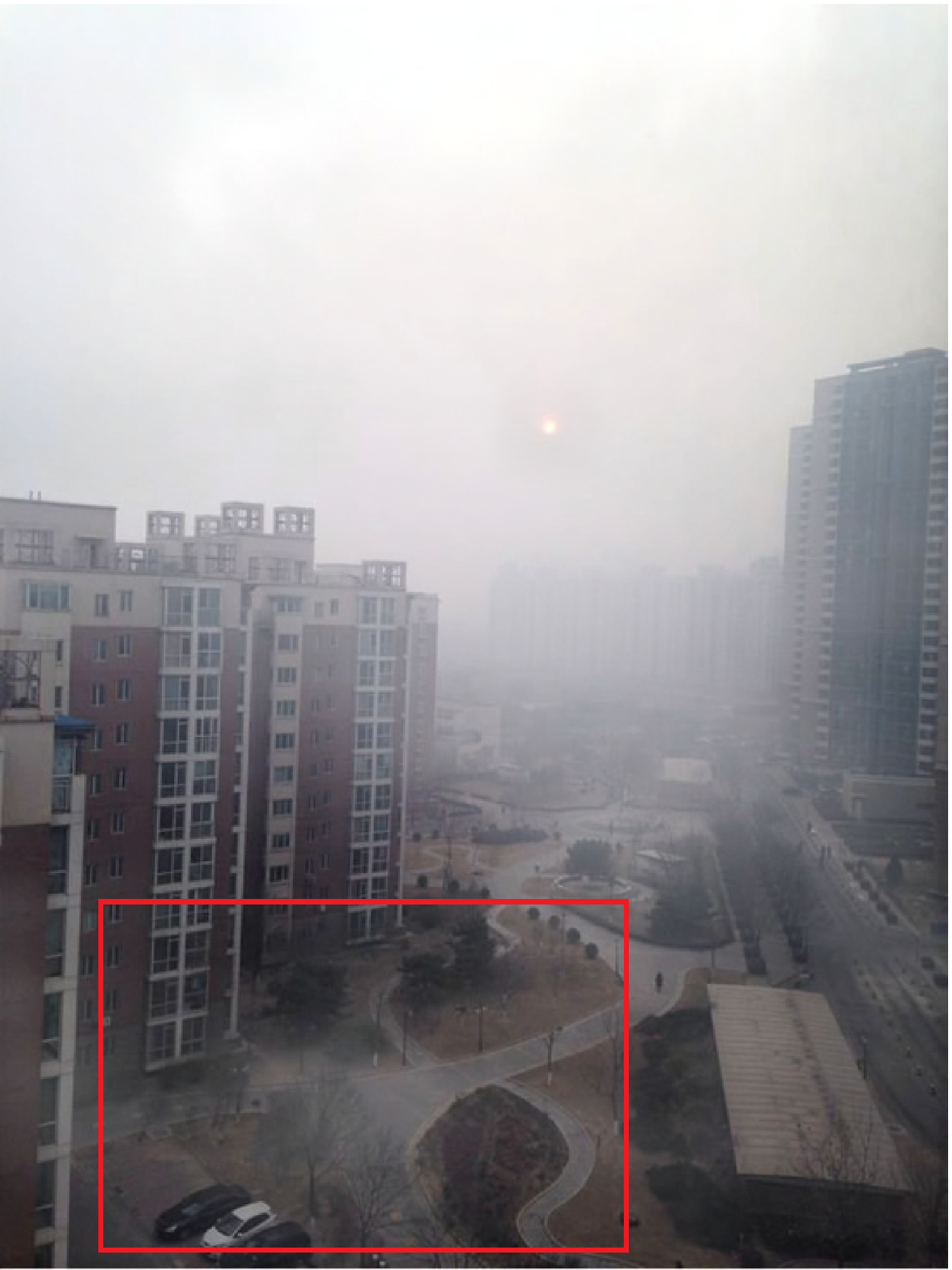}
    \end{minipage}
}
\subfigure[FFA-Net\cite{qin2020ffa}]
{
    \begin{minipage}[b]{.15\linewidth}
        \centering
        \includegraphics[scale=0.147]{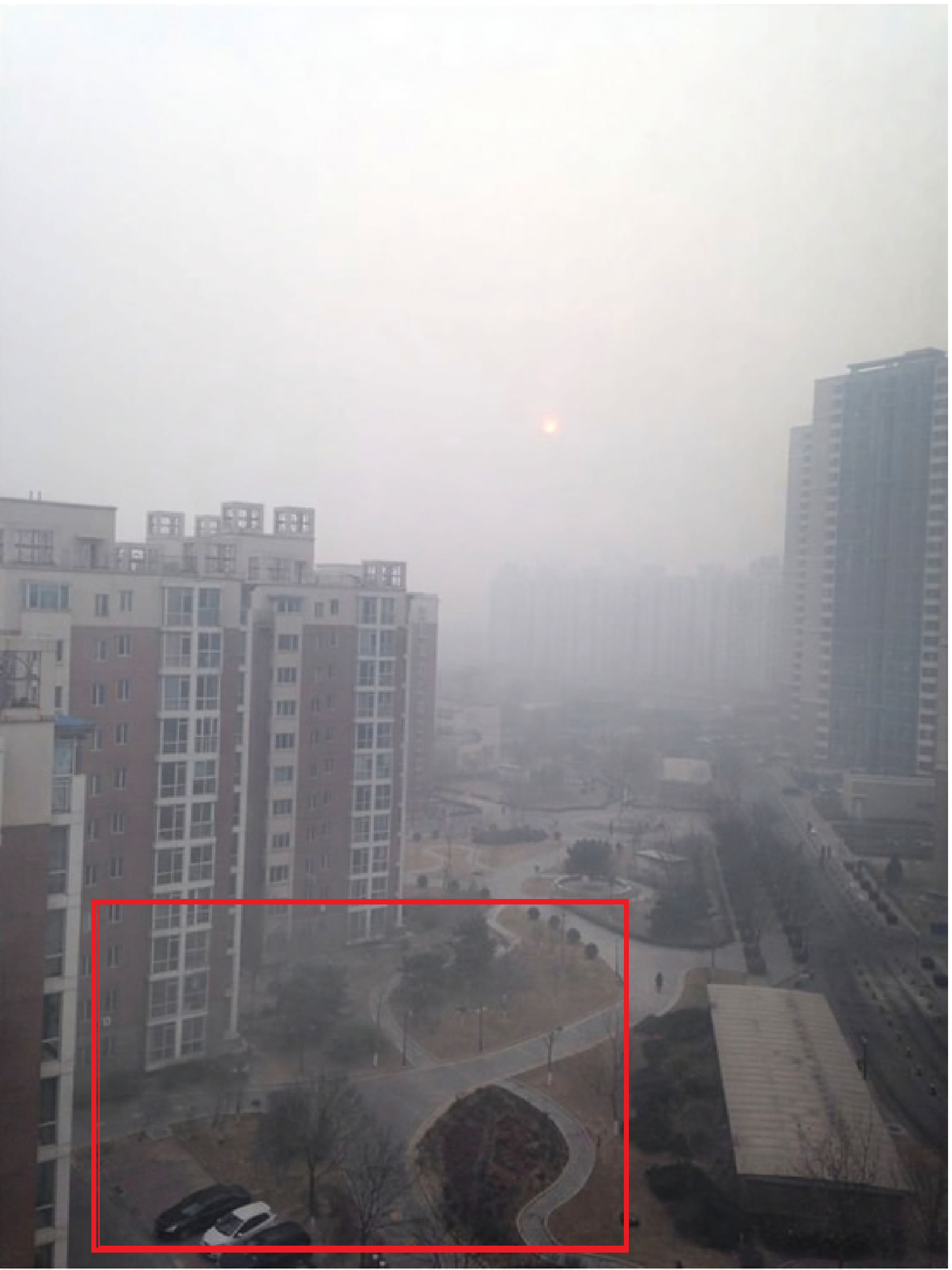}
    \end{minipage}
}
\subfigure[ours]
{
    \begin{minipage}[b]{.15\linewidth}
        \centering
        \includegraphics[scale=0.147]{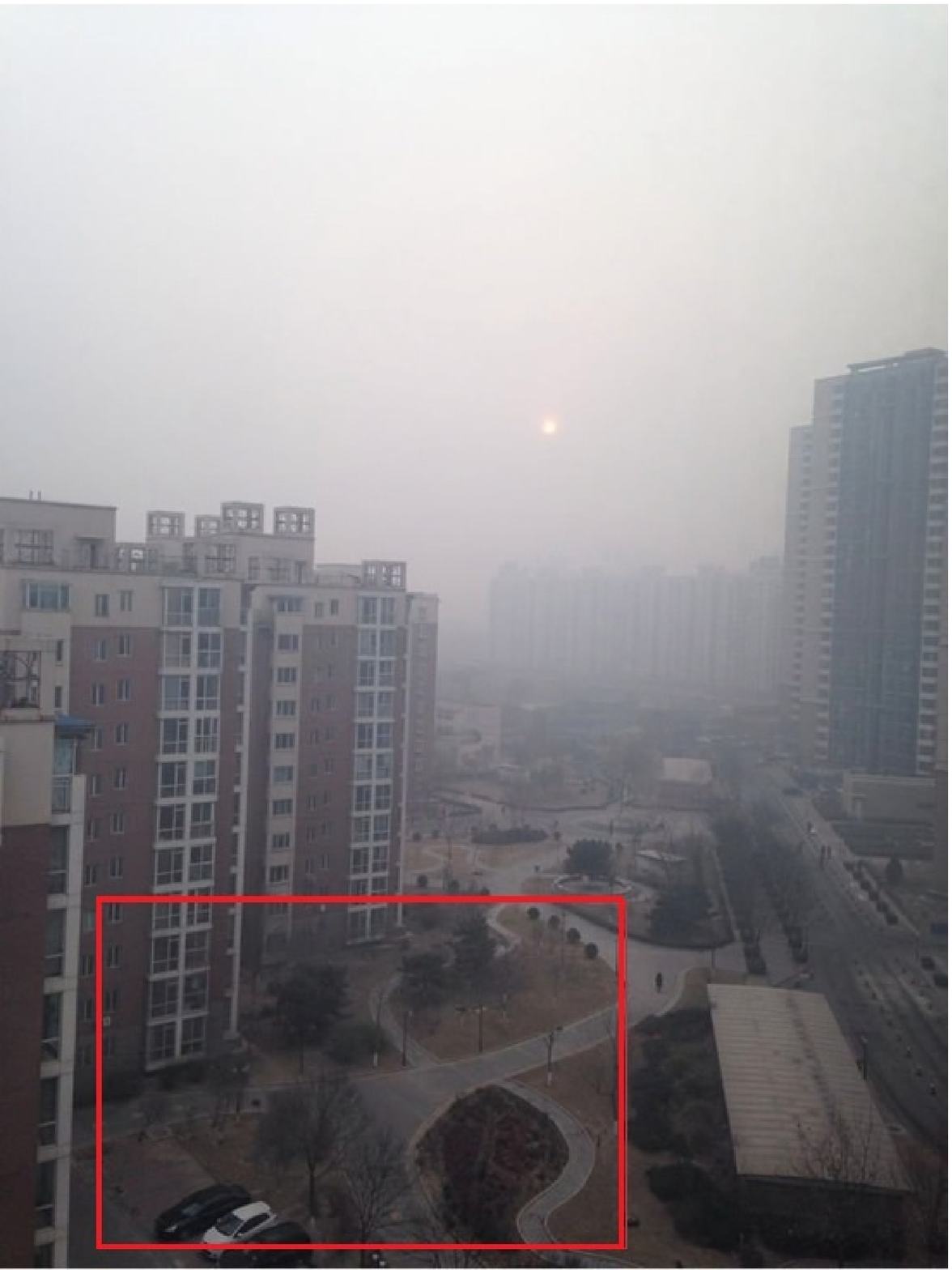}
    \end{minipage}
}
\subfigure[GT]
{
    \begin{minipage}[b]{.15\linewidth}
        \centering
        \includegraphics[scale=0.147]{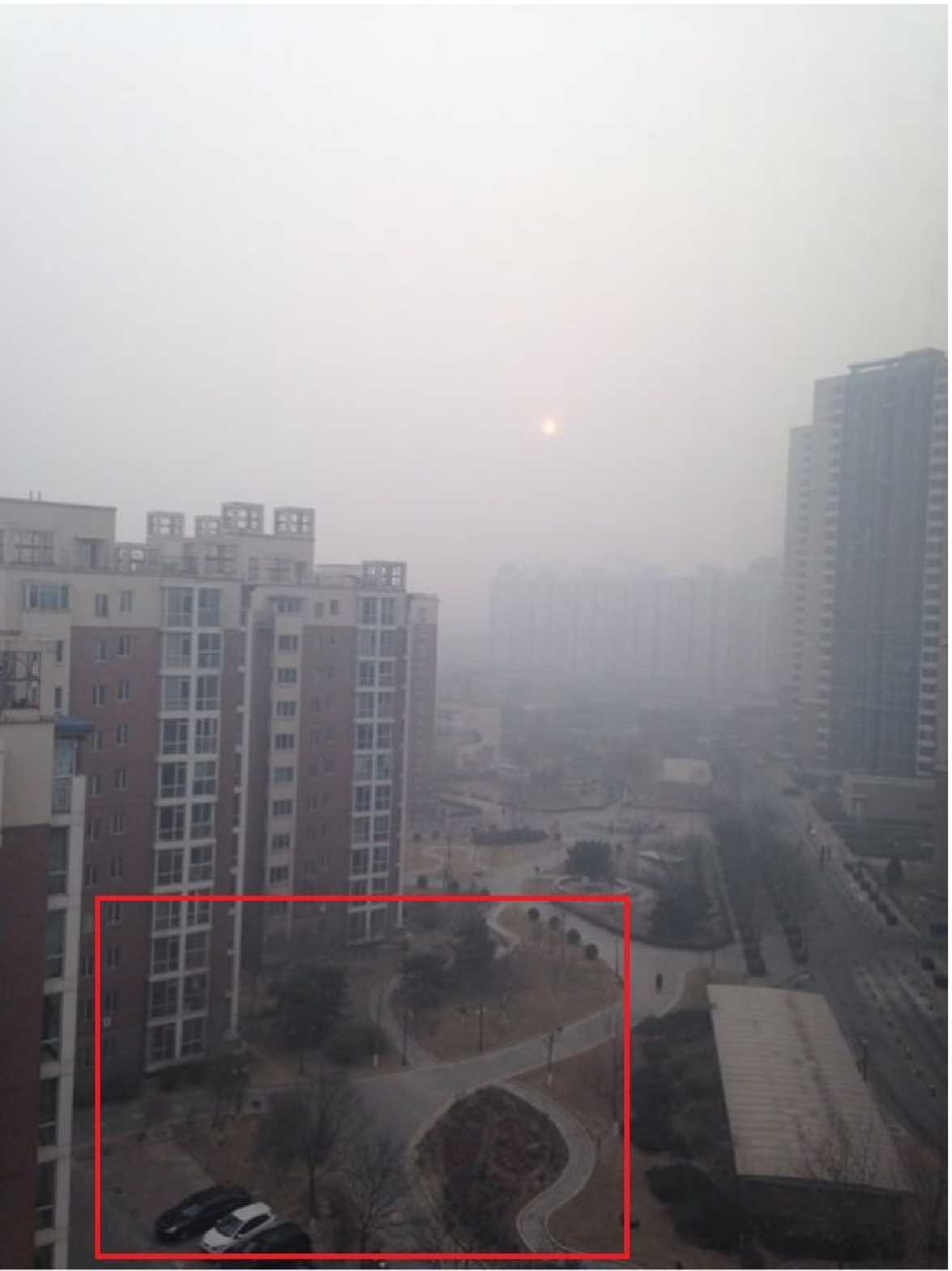}
    \end{minipage}
}
\caption{Qualitative comparisons on RESIDE-OUT dataset. Zoom in for the best view.}
\label{Figure 6}
\end{figure*}

\subsection{Comparison with state-of-the-art methods }
\textbf{Quantitative Analysis}: We compared the performance of MixDehazeNet with previous state-of-the-art methods, and the results are shown in Table \ref{Table 1}. Our model outperforms all previous methods in all three datasets. In the RESIDE-IN dataset, our MixDehazeNet-L model is the first method to exceed 42dB PSNR, and it outperforms all previous state-of-the-art methods with a large margin in terms of both PSNR and SSIM. In the RESIDE-OUT and RESIDE-6k datatsets, our MixDehazeNet-L model also outperforms all previous state-of-the-art methods in terms of PSNR and SSIM. Our different model variants showed excellent performance, with PSNR increasing with the number of mix structure blocks. We believe that our methods can be adapted to different types of computer vision tasks. MixDehazeNet-S can be applied to real-time image dehazing, while MixDehazeNet-L can be applied to image dehazing tasks with higher definition requirements.

\textbf{Qualitative Analysis}: Figure \ref{Figure 5} shows the visual results of our MixDehazeNet compared to previous state-of-the-art models on RESIDE-IN dataset. The restored images produced by DCP \cite{DCPDN}, GridDehazeNet \cite{Griddehazenet} and FFA-Net \cite{qin2020ffa} all contain different levels of artifacts, which reduced the clarity of the images. In contrast, the images restored by our model are the clearest and closest to the ground truth without any artifacts. Similarly, Figure \ref{Figure 6} shows the visual results of our MixDehazeNet compared to previous state-of-the-art models on RESIDE-OUT dataset. Since the prior knowledge is not satisfied, the images recovered by DCP \cite{DCPDN} have color distortion. Both GridDehazeNet \cite{Griddehazenet} and FFA-net \cite{qin2020ffa} have more haze residue and the distribution of residual haze in the restored image is uneven. In comparison, the restored images of our model are clearer with less haze residue and sharper edge contours and are closest to the ground truth. The red rectangles in the figure highlight the differences in the details of the restored images produced by each model.

\begin{table}[t]
\renewcommand\arraystretch{1.5}
\centering
\caption{\centering Model Architecture Detailed.}
\label{Table 2}
\setlength{\tabcolsep}{0.8mm}{
\begin{tabular}{c|c|c}
\hline
Model Name& Num. of Blocks& Embedding Dims\\
\hline
% MixDehazeNet-T& [1 , 1 , 2 , 1 , 1]& [24 , 48 , 96 , 48 , 24]\\
MixDehazeNet-S& [2 , 2 , 4 , 2 , 2]& [24 , 48 , 96 , 48 , 24]\\
MixDehazeNet-B& [4 , 4 , 8 , 4 , 4]& [24 , 48 , 96 , 48 , 24]\\
MixDehazeNet-L& [8 , 8 , 16 , 8 , 8]& [24 , 48 , 96 , 48 , 24]\\
\hline
\end{tabular}}
\end{table}

\subsection{Ablation Study}
The ablation experiments were conducted on the MixDehazeNet-S model to understand the role of each proposed module. We start with MixDehazeNet-S-Base. (1) Each Mix structure block in the MixDehazeNet-S-Base retains DWDConv19 (remove DWDConv7 and DWDConv13) in the Multi-Scale Parallel Large Convolution Kernel (MSPLCK) and retains channel attention (remove simple pixel attention and pixel attention) in the Enhanced Parallel Attention (EPA). (2) We only recover MSPLCK in MixDehazeNet-S-Base. (3) We only recover EPA in MixDehazeNet-S-Base. (4) We recover MSPLCK and EPA in MixDehazeNet-S-Base. And then, we add contrastive loss (CR) to the entire model. All ablation model training configuration corresponds to MixDehazeNet-S and experimented on the RESIDE-IN dataset. The results of the ablation study can be found in Table \ref{Table 3}. MSPLCK can increase 1.76 dB PSNR compared to MixDehazeNet-S-Base. EPA can increase 1.89 dB PSNR compared to MixDehazeNet-S-Base. Mix Structure block which combines MSPLCK and EPA can increase 3.96 dB PSNR compared to MixDehazeNet-S-Base. The results indicate that each proposed module can improve the dehazing performance of the model.

\begin{table}[t b p]
\renewcommand\arraystretch{1.5}
\centering
\caption{\centering Ablation Study on the RESIDE-IN dataset.}
\label{Table 3}
\setlength{\tabcolsep}{4.5mm}{
\begin{tabular}{c|c|c}
\hline
Method & PSNR & SSIM\\
\hline
MixDehazeNet-S-Base&34.39 &0.987 \\
+MSPLCK &36.15 &0.990 \\
+EPA &36.28 &0.991 \\
+MSPLCK+EPA& 38.35 &  0.992\\
+MSPLCK+EPA+CR& 39.47 & 0.995 \\
\hline
\end{tabular}} 
\end{table}

To further verify the role of the two proposed modules, two sets of ablation experiments were conducted. To speed up the experiments, a tiny version of the model was proposed with a simplified architecture detailed in Table \ref{Table 4} and the training epoch was reduced to 400. The learning rate was set from $4 \times 10^{-4}$ to $4 \times 10^{-6}$ with the cosine annealing strategy. 

\textbf{Multi-Scale Parallel Large Convolution Kernel}: In order to verify the multi-scale characteristics and large receptive fields of MSPLCK, we conducted three sets of comparison experiments using same-scale parallel large convolution kernels. Firstly, we replaced the multi-scale dilated convolution in MSPLCK to Three Parallel DWDConv7 which dilated convolution kernel size = 7 and it is $3\times3$ depth-wise dilated convolution with dilation rate 3. Next, we replaced the multi-scale dilated convolution in MSPLCK with Three Parallel DWDConv13 which dilated convolution kernel size = 13 and it is $5\times5$ depth-wise dilated convolution with dilation rate 3. Finally, we replaced the multi-scale dilated convolution in MSPLCK with Three Parallel DWDConv19 which dilated convolution kernel size = 19 and it is $7\times7$ depth-wise dilated convolution with dilation rate 3. Table \ref{Table 5} shows experimental results, MSPLCK which has multi-scale characteristics shows a better effect than other same-scale parallel large convolution kernels. And the larger receptive fields of the convolution kernel resulted in a better haze removal effect.

\textbf{Enhanced Parallel Attention}: In order to verify whether parallel Attention is more suitable for image dehazing, we conducted two sets of comparison experiments to compare the effects between serial Attention and parallel attention. First, we replaced the three parallel attention in EPA to channel attention (CA) and pixel attention (PA) in serial, and then we replaced the three parallel attention in EPA with simple pixel attention (SPA), channel attention (CA) and pixel attention (PA) in serial. Table \ref{Table 6} shows experimental results, indicating that the parallel attention mechanism is more suitable for image dehazing than the serial attention mechanism. 
\begin{table}
\renewcommand\arraystretch{1.5}
\centering
\caption{\centering Model Architecture Detailed.}
\label{Table 4}
\setlength{\tabcolsep}{0.8mm}{
\begin{tabular}{c|c|c}
\hline
Model Name& Num. of Blocks& Embedding Dims\\
\hline
MixDehazeNet-T& [1 , 1 , 2 , 1 , 1]& [24 , 48 , 96 , 48 , 24]\\
\hline
\end{tabular}}
\end{table}

\begin{table}[t b p]
\renewcommand\arraystretch{1.5}
\centering
\caption{\centering Multi-Scale Parallel Large Convolution kernel ablation study on the RESIDE-IN dataset.}
\label{Table 5}
\setlength{\tabcolsep}{3.5mm}{
\begin{tabular}{c|c|c}
\hline
Method & PSNR & SSIM\\
\hline
% Three Parallel DWDConv7(Epoch 300)& 33.69 & 0.986\\
% Three Parallel DWDConv13(Epoch 300)& 33.82 & 0.986\\
% Three Parallel DWDConv19(Epoch 300)&34.01 &0.986 \\
Three Parallel DWDConv7& 35.23 & 0.989\\
Three Parallel DWDConv13& 35.67 & 0.990 \\
Three Parallel DWDConv19&35.71 &0.990 \\
MSPLCK&35.94 &0.990 \\
\hline
\end{tabular}}
\end{table}

\begin{table}[t b p]
\renewcommand\arraystretch{1.5}
\centering
\caption{\centering Enhanced Parallel Attention ablation study on the RESIDE-IN dataset.}
\label{Table 6}
\setlength{\tabcolsep}{4.5mm}{
\begin{tabular}{c|c|c}
\hline
Method & PSNR & SSIM\\
\hline
%CA PA in Serial(Epoch 300 四卡)&34.20&0.986 \\
CA PA in Serial&34.95&0.988 \\
SPA CA PA in Serial&34.25&0.987 \\
EPA&35.94 &0.990 \\
%EPA(Epoch 300 四卡)&34.25 &0.987 \\
%EPA(Epoch 500 四卡)&34.41 &0.987 \\
\hline
\end{tabular}
}
\end{table}

\subsection{Inference Time}
In Table \ref{Table 1}, we also compared the inference speed of our model to previous state-of-the-art models. Our model performs significantly better while maintaining a similar inference time. For example, on the RESIDE-IN dataset, MixDehazeNet-S and MSBDN \cite{MSBDN} have a similar inference time of approximately 14 ms, but MixDehazeNet-S has increased 5.8dB PSNR compared to MSBDN. MixDehazeNet-B has a similar inference time of about 28 ms with AECR-Net \cite{AECR-net} and PMNet \cite{PMNet}, but it has increased 3.37dB PSNR and 2.49dB PSNR compared with AECR-Net and PMNet respectively.  MixDehazeNet-L and FFA-Net \cite{qin2020ffa} have a similar inference time of 56 ms, but MixDehazeNet-L has increased 6.23dB PSNR compared with FFA-Net.
%-------------------------------------------------------------------------
\section{Conclusion}
In this paper, we propose MixDehazeNet, which contains mix structure block consisting of a multi-scale parallel large convolution kernel module and an enhanced parallel attention module. The multi-scale parallel large convolution kernel to achieve multi-scale large receptive fields. The enhanced parallel attention efficiently deals with uneven hazy distribution and allow useful features to pass through the backbone. To our best knowledge, our method is the first to exceed 42dB PSNR in the RESIDE-IN dataset.

% \section{Acknowledgements}
% This work was supported by the National key R\&D Program(2021YFB2501104), the National Natural Science Foundation of China under Grant(62206204).

% {\small
% \bibliographystyle{ieee_fullname}
% \bibliography{egbib}
% }

\end{document}